%% file: main.tex
\pdfoutput=1
\PassOptionsToPackage{table}{xcolor}

\documentclass[11pt]{article}
\usepackage{authblk}
\usepackage{float}
\usepackage{ACL2023}
\usepackage{multirow}
\usepackage{times}
\usepackage{latexsym}
\usepackage{amsmath,amssymb}
\usepackage[T1]{fontenc}
\usepackage{booktabs}
\usepackage{enumitem}  
\usepackage{xcolor}
\usepackage{siunitx}  

\usepackage[utf8]{inputenc}

\usepackage{microtype}

\usepackage[full]{textcomp}

\usepackage{inconsolata}
\usepackage{tikz}
\usepackage{tcolorbox} 
\usepackage[normalem]{ulem}  
\usepackage{makecell}  
\usepackage{pifont}  
\usepackage{adjustbox}  
\usepackage{footmisc}
\usepackage{stix}

\newif\ifshowcomments
\showcommentstrue
\showcommentsfalse

\ifshowcomments

\newcommand{\authorcomment}[3]{\textcolor{#1}{[\textsf{#2} -- #3]}}
\newcommand{\arie}[1]{\authorcomment{purple}{#1}{Arie}}
\newcommand{\shmulik}[1]{\authorcomment{blue}{#1}{Shmulik}}
\newcommand{\OS}[1]{\authorcomment{cyan}{#1}{Ori}}
\newcommand{\eviatar}[1]{\authorcomment{red}{#1}{Eviatar}}
\newcommand{\ido}[1]{\authorcomment{green}{#1}{Ido}}

\def\XXX#1{\textcolor{red}{XXX #1}}
\def\repl#1#2{\textcolor{red}{XXX \sout{#1}}\textcolor{blue}{\uline{#2}}}

\else

\newcommand{\authorcomment}[3]{}
\newcommand{\arie}[1]{}
\newcommand{\shmulik}[1]{}
\newcommand{\eviatar}[1]{}
\newcommand{\OS}[1]{}
\newcommand{\ido}[1]{}

\def\XXX#1{}
\def\repl#1#2{#2}

\fi

\newcommand{\ourdataset}{\textsc{NatConfQA}}
\newcommand{\ourtask}{\textsc{ConfMAQA}}
\newcommand{\climatefever}{\textsc{climate-fever}}
\newcommand{\healthver}{\textsc{HealthVer}}

\usepackage{xcolor}

\newcommand{\cmark}{\ding{51}}%
\newcommand{\xmark}{\ding{55}}%

\definecolor{pastelRed}   {HTML}{DA2C43}  
\definecolor{pastelRed2}  {HTML}{780606}  
\definecolor{pastelGreen} {HTML}{00BB77}  
\definecolor{pastelGreen2}{HTML}{00E893}  
\definecolor{pastelGrey}  {HTML}{6D8196}  
\definecolor{pastelGrey}  {HTML}{6D8196}  

\title{Consensus or Conflict?\\ Fine-Grained Evaluation of Conflicting Answers in Question-Answering}

\usepackage{authblk}  

\author[1]{\bf Eviatar Nachshoni}
\author[1,2]{\bf Arie Cattan}
\author[1]{\bf Shmuel Amar}
\author[3]{\bf Ori Shapira}
\author[1,2]{\bf Ido Dagan}  

{
\makeatletter
\renewcommand\AB@affilsepx{~~~~~~ \protect\Affilfont}
\makeatother

\affil[1]{Bar-Ilan University}
\affil[2]{Google Research}
\affil[3]{OriginAI}
}

\affil[ ]{}

\affil[ ]{\tt \{eviatarn, shmulikamar\}@gmail.com ~ cattana@google.com}
\affil[ ]{\tt obspp18@gmail.com ~ dagan@cs.biu.ac.il}

\begin{document}
\maketitle
\input{00abstract}
\input{Figures/ConflictExample}

\input{01introduction_arie}
\input{03task_definition_arie}

\input{04dataset_methoology}

\input{05experimental_setting}
\input{06Results}
\input{02related_work_arie}
\input{07conclusions}

\input{10limitations}
\input{11Acknowledgements}

\bibliographystyle{acl_natbib}
\bibliography{anthology,custom}

\input{90Appendix}

\end{document}

%% file: 00abstract.tex
\begin{abstract}

Large Language Models (LLMs) have demonstrated strong performance in question answering (QA) tasks. However, Multi-Answer Question Answering (MAQA), where a question may have several valid answers, remains challenging. Traditional QA settings often assume consistency across evidences, but MAQA can involve conflicting answers. Constructing datasets that reflect such conflicts is costly and labor-intensive, while existing benchmarks often rely on synthetic data, restrict the task to yes/no questions, or apply unverified automated annotation. To advance research in this area, we extend the conflict-aware MAQA setting to require models not only to identify all valid answers, but also to detect specific conflicting \textit{answer pairs}, if any. To support this task, we introduce a novel cost-effective methodology for leveraging fact-checking datasets to construct \ourdataset{}, a new benchmark for realistic, conflict-aware MAQA, enriched with detailed conflict labels, for all answer pairs. We evaluate eight high-end LLMs on \ourdataset{}, revealing their fragility in handling \repl{}{various types of} conflicts and the flawed strategies they employ to resolve them.\footnote{\ourdataset{} is publicly available at \url{https://github.com/EN555/ContraQA}.}

\end{abstract}

%% file: Figures/ConflictExample.tex
\newcommand{\udashA}[1]{%
    \tikz[baseline=(todotted.base)]{
        \node[inner sep=1pt,outer sep=0pt] (todotted) {#1};
        \draw[line width=1pt, dashed] (todotted.south west) -- (todotted.south east);
    }%
}%

\newcommand{\udashB}[1]{%
    \tikz[baseline=(todotted.base)]{%
        \node[inner sep=1pt,outer sep=0pt] (todotted) {#1};%
        \draw[line width=1pt, dash pattern=on 2pt off 1pt]
            ([yshift=-1pt]todotted.south west) -- ([yshift=-1pt]todotted.south east);%
    }%
}%

\newcommand{\udashC}[1]{%
    \tikz[baseline=(todotted.base)]{%
        \node[inner sep=1pt,outer sep=0pt] (todotted) {#1};%
        \draw[line width=1pt, dash pattern=on 4pt off 4pt]
            ([yshift=-1pt]todotted.south west) -- ([yshift=-1pt]todotted.south east);%
    }%
}%

\newcommand{\colorGreenDash}[1]{\textcolor{pastelGreen}{\udashA{#1}}}
\newcommand{\colorRedDash}[1]{\textcolor{pastelRed}{\udashB{#1}}}
\newcommand{\colorGrayDash}[1]{\textcolor{pastelGrey}{\udashC{#1}}}

\begin{figure}[!t]
  \centering
  \resizebox{\linewidth}{!}{%
\includegraphics[page=1, width=\linewidth, trim={0 276 1172 0}, clip]{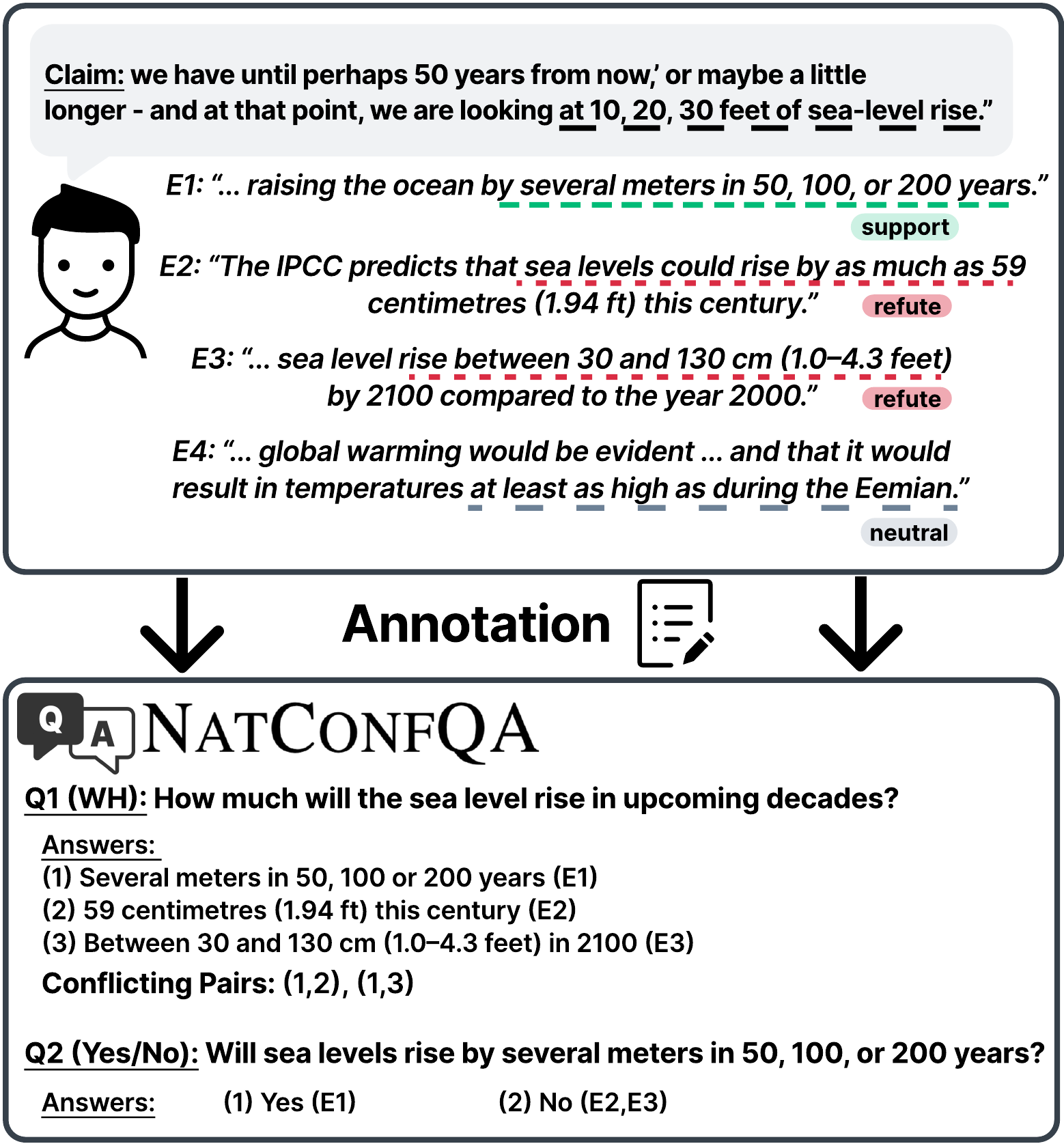}%
}
\caption{Deriving a conflict-aware MAQA instance from a fact-chcecking instance.
The source instance is composed of a claim with \colorGreenDash{supporting}, \colorRedDash{refuting} and \colorGrayDash{neutral} evidence.
Annotators then create WH and Yes/No questions to surface these conflicts and label \textit{conflicting} answer pairs.}
  \label{fig:fact-checking-annotation}
\end{figure}


%% file: 01introduction_arie.tex
\section{Introduction}

Recent advances in Large Language Models (LLM) \citep{fischer2024questionlargelanguagemodels, openai2024gpt4technicalreport} have led to substantial performance improvement in various tasks, including Question Answering (QA) with one or multiple correct answers~\citep{voorhees2004trec2003}. 
Although previous work on multi-answer QA (MAQA) largely assumes that the different answers are mutually consistent and complementary~\citep{zhong2022romqabenchmarkrobustmultievidence, amouyal-etal-2023-qampari}, realistically, questions can be controversial and lack a definitive answer. In such cases, models should not only generate a response that incorporates several answers, but also detect the conflicts and communicate them to the reader. For example, when asked \emph{``What is the effect of aspartame?''}, a comprehensive response should aggregate various effects reported in the available sources, while explicitly distinguishing between effects with consensus and those that remain contested or under debate.

Collecting QA instances with naturally occurring contradictory answers is challenging, as knowledge conflicts are not always prevalent in arbitrary texts. As a result, only a few datasets handle knowledge conflicts while exhibiting some limitations, such as relying on LLMs to inject misinformation into other reliable texts, or focusing only on Yes/No questions (Section~\ref{sec:related_work}). Furthermore, although recent benchmarks aim to evaluate whether the \textit{entire} response acknowledges the debatable nature of the question~\citep{xu2024debateqaevaluatingquestionanswering, hou2024wikicontradictbenchmarkevaluatingllms}, they do not assess whether the response accurately reflects \emph{which} answers are subject to disagreements (e.g., whether aspartame increases cancer risk).

To foster research on this important challenge and to enable fine-grained evaluation, we create \ourdataset{}, the first conflict-aware MAQA dataset with annotations labeling individual answer pairs that are in conflict. To collect \ourdataset{}, we first leverage standard fact-checking datasets to identify sources with naturally occurring disagreements. Then, we ask human annotators to write various QA pairs and to label the relationship between pairs of conflicting answers. Finally, we verify the annotations for quality assurance. \autoref{fig:fact-checking-annotation} illustrates our general annotation scheme. \ourdataset{} is a high-quality dataset that covers Yes/No and WH- questions, and includes instances, based on reliable sources,
with a mix of conflicting and non-conflicting answers for the same question.

We evaluate the performance of eight LLMs on \ourdataset{}, including open-source and proprietary models, measuring both answer quality and conflict identification. In terms of answer quality, we show that while models generally achieve high precision, they fail to output all correct answers. Furthermore, our fine-grained evaluation of conflict identification reveals that models struggle to distinguish between conflicting and non-conflicting answer pairs. Further analysis of the model failures reveal insightful error patterns: models evade exposing conflicts by selecting a single answer, erroneously attempting to reconcile contradictory information, or refraining from answering the question altogether. Taken together, 
our work uncovers the behavior of strong LLMs when confronted with conflicting information, while providing suitable methodologies and data to investigate these challenges in future research.

%% file: 03task_definition_arie.tex
\section{Conflict-Aware Multi-Answer QA}
\label{sec:task-defintion}

\input{Tables/ExampleInstance}

The \textit{Conflict-Aware Multi-Answer QA} task is an extension of the traditional Multi-Answer QA \cite{min-etal-2020-ambigqa, amouyal-etal-2023-qampari} task, that considers potential conflicts between the different answers. We extend recent work in conflict-aware QA \cite{xu2024debateqaevaluatingquestionanswering}, which either focused on binary conflicting answers \citep{hou2024wikicontradictbenchmarkevaluatingllms} or addressed multiple answers without indicating which pairs conflict \citep{jiayang2024econdetectionresolutionevidence}. Our generalized task formulation supports two or more answers per question and, importantly, pinpoints the answer pairs that are in conflict.

Given a question $q$ and a set of candidate passages $P = \{p_1, \dots, p_n\}$, the task is to generate a free-text response $y$ that satisfies two main requirements. First, similar to the traditional MAQA task, $y$ should incorporate all answers that appear in reference $A = \{a_1, \dots, a_m\}$.
Second, the response $y$ should indicate all conflicts, if any, between the answers within it. We assume that the response $y$ is in natural language, as typically generated by large language models (LLMs), and not necessarily in a structured format.

For example, consider the question \textit{``What climate degree change is caused by greenhouse gases?"}. The different answers \textit{0.45°C} and \textit{0.8°C} cannot be simultaneously true, hence an ideal response should present both answers through contrastive language (e.g., by using the word ``however'', both answers can be communicated, and the conflict is established). In contrast, for the question \textit{``Which domestic pets can potentially test positive for SARS-CoV-2?"}, the answers \textit{dogs}, \textit{cats}, and \textit{ferrets} are non-conflicting and an ideal response should enumerate the different answers cohesively.

Since a question may elicit a mixture of conflicting and non-conflicting answers, we want to determine whether the model response $y$ accurately reflects the conflict or non-conflict relations between the different answer pairs. 
Formally, for reference answers  $A$ and a respective set $C=\bigl\{\{a_i,a_j\}\mid a_i\text{ and }a_j\text{ conflict}\bigr\}$, that lists the pairs of conflicting answers in $A$, the objective of response $y$ is to accurately incorporate $A$ and $C$.

In accordance with the task definition, we define two evaluation criteria, adopted from related tasks \citep{min-etal-2020-ambigqa, hou2024wikicontradictbenchmarkevaluatingllms}: \textit{Answer Quality}, measuring how well the model covers the set of correct answers; and \textit{Conflict Identification}, assessing the model’s ability to correctly identify conflicting answer pairs. We propose metrics for the evaluation criteria in \S\ref{subsec:evaluation-metrics}.

%% file: Tables/ExampleInstance.tex

%% file: 04dataset_methoology.tex
\section{Creating the \ourdataset{} Dataset}
\label{sec:annotation-methodolgy}

Obtaining annotations for a conflict-aware MAQA dataset is challenging, because informational conflicts are infrequent in arbitrary sources from which answers can be collected.


In this work, we approach this challenge by leveraging existing fact-checking datasets, a well-studied field with many datasets, which were annotated with large manual effort \cite{thorne-etal-2018-fever, sarrouti-etal-2021-evidence-based}.
Some fact-checking datasets contain claims for which some pieces of evidence refute the claim, while others support it,
as exemplified in \autoref{fig:fact-checking-annotation}. 
This structure makes them particularly valuable for our task, as they naturally capture conflicting evidence.

In this section, we describe our methodology for converting and annotating existing fact-checking datasets into a conflict-aware MAQA dataset (\S\ref{sec:fact-cehcking-to-qa} and \S\ref{subsec:annotation-process}). We then describe the resulting new \ourdataset{} dataset (\S\ref{sec:instance_structure}) finally, we examine its quality (\S\ref{sec:data_quality}).


\subsection{From Fact-Checking to Conflict-Aware Multi-Answer QA}
\label{sec:fact-cehcking-to-qa}

The typical structure of an instance in a fact-checking dataset is a triplet \( (c, e, \ell) \), where \(c\) is a claim, \(e\) is a piece of evidence associated with that claim, and \(\ell\in\{\mathit{refute},\mathit{support},\mathit{neutral}\}\) indicates the entailment relation between the evidence $e$ and the claim $c$ \cite{fever-2021-fact,sarrouti-etal-2021-evidence-based}. 
In the \climatefever{} \citep{diggelmann2021climatefeverdatasetverificationrealworld} and \healthver{} \citep{sarrouti-etal-2021-evidence-based} datasets, the same claim \(c\) may appear in multiple triplets \((c,e_i,\ell_i)\), where it might be supported by some pieces of evidence, while others refute it. The co-occurrence of both supporting and refuting evidence for the same claim typically indicates the presence of an underlying factual conflict among the pieces of evidence (see App.~\ref{sec:fact-checking-datasets} for details).

Our goal is to leverage the above fact-checking datasets in order to create a conflict-aware MAQA dataset. An instance should contain a question $q$ and a set of respective answers $A = \{a_1, \dots, a_k\}$ that includes at least one pair of conflicting answers. In reality, however, not all questions have conflicting answers. Hence, we would also like to include instances where $A$ has only non-conflicting answers. Such a subset is useful as a control set when assessing models' performance in the conflict-aware MAQA task (\S\ref{sec:results}).
Beyond accommodating conflicting and non-conflicting answer sets, it is also important to support variation in question form, i.e., WH-questions versus Yes/No questions.

Overall, we gather two sets of instances from fact-checking datasets: (1) \emph{conflicting} instances, that include both supporting and refuting pieces of evidence, and (2) \emph{non-conflicting} instances, that include at least one supporting evidence and no refuting evidence. We next describe our process for converting these fact-checking instances into ones for the conflict-aware MAQA task.

\subsection{Dataset Preparation}
\label{subsec:annotation-process}

\paragraph{Initial fact-checking data.}
We first gathered the two sets of fact-checking instances on which the annotation process is conducted. For the \textit{conflicting} set, we automatically iterated over the \climatefever{} and \healthver{} instances (see \autoref{sec:fact-checking-datasets} for pre-processing details), and grouped those with the same claim, that have conflicting evidence (at least one triplet with $\mathit{support}$ and at least one with $\mathit{refute}$), resulting in 188 groups. For the \textit{non-conflicting} set, we collected \shmulik{How much?} several hundred groups of claims with only supporting or neutral evidence.

The fact-checking datasets supply evidence at the sentence level, sourced from Wikipedia articles in \climatefever{} and CORD‐19 abstracts in \healthver{}. In a realistic situation, especially in the QA setting, the source texts on which the task is performed are typically longer passages. Therefore, for each evidence sentence used from the fact-verification datasets, we retrieved the complete passage containing that sentence (details in \autoref{subsec:passage-extraction}).\footnote{Note that a passage can contain more than one evidence sentences if they come from the same source passage.} In summary, each instance in the two fact-checking sets is composed of a claim and several pieces of evidence within their passages.


\repl{We applied an LLM \citep[\texttt{gpt-4o-mini};][]{openai2024gpt4o} to generate MAQA instances based on the two fact-checking sets (see \autoref{fig:suggestion-for-qa} in the appendix for the prompts). We found that the LLM-generated question-answer pairs, especially those intended to expose conflicts, lacked sufficient quality.\footnote{We observed that \repl{ the model}{even top-performing models} often mark\repl{ed}{} instances as containing conflicts even when none exist\repl{ed}{}, or, in instances where conflicts were present, \repl{it}{they} fail\repl{ed}{} to clearly reflect them in the generated question.}
We therefore hired two annotators to create the new instances, which were assisted by LLM-generated responses serving as intermediate candidates.}{}

\paragraph{Manual annotation process.}

The two annotators were first provided each with half of the \textit{conflicting} fact-checking instances.\footnote{We observed that even top-performing models often mark instances as containing conflicts even when none exist, or, in instances where conflicts were present, they fail to clearly reflect them in the generated question.} Each instance is composed of a claim and several pieces of evidence (some conflicting)\repl{, and an intermediate LLM-generated suggestion}{}. The annotators then followed the following procedure (more details in \autoref{fig:app-annotation-guidelines}). (1) Contradiction detection: confirm whether the supporting and refuting evidence indeed conflict with each other. \repl{This resulted in 89 of 188 (47\%) suitable instances.}{}\shmulik{I dont think we should remove this sentence.} (2) WH-question formation: for each instance, write a WH-question based on the claim and evidence, aiming to elicit the core information and potential conflict. Then for each question, write out a list of answers based on the evidence, and link the evidence to the answer. (3) Label answer pairs: label each pair of answers as \textit{conflicting} or \textit{non-conflicting}. (4) Yes/no question formation: for each \repl{pair of conflicting answers}{WH-question and its corresponding answer} from the previous steps, if possible, formulate a yes/no question and link the supporting evidence for the yes and no responses.

The annotators then repeated the process on the \textit{non-conflicting} fact-checking set, skipping steps 1 and 3. Also, step 2 requires questions and answers with only non-conflicting evidence, and step 4 is conducted for all answers from step 2 in this round.  \repl{(each WH-question and its corresponding answer are paraphrased into additional yes/no instances).}{}
\repl{}{See \autoref{sec:annotation-tool} for details regarding annotators and the annotation tool.}

\subsection{The \ourdataset{} Dataset}\label{sec:instance_structure}

\repl{}{Overall, the annotation process yielded 269 conflicting instances, of which 89 were WH-questions, and 408 non-conflicting instances.}
Each instance in \ourdataset{} is represented as a tuple \( (q, P, A, C \)), where \( q \) is the question, \( P = \{p_1, \dots, p_n\} \) is the set of relevant passages, \( A = \{a_1, \dots, a_m\} \) is the set of answers, and \(C=\bigl\{\{a_i,a_j\}\mid a_i\text{ and }a_j\text{ conflict}\bigr\}\) contains all annotated pairs of conflicting answers.\footnote{Each answer is also linked to the passage(s) containing its respective evidence.}

\repl{\autoref{tab:support_conflict_distribution_merged} presents statistics for \ourdataset{},\footnote{For randomly sampled instances from the dataset, see \autoref{tab:sample-instances}.} including number of questions, and number of passages, answers and conflicts per question. The dataset comprises two subsets, one with only supporting answers (titled \textit{No-conflict} in the table) and another with conflicting answers (titled \textit{Conflict}). For both types, there are WH-questions as well as Yes/No questions.}{}

\subsection{Dataset Quality}
\label{sec:data_quality}


\shmulik{TODO: We must move table statistics to Appendix}

\paragraph{Data validation.}

A high-quality dataset should align with the objectives of our task and the evaluation criteria (\S\ref{sec:task-defintion}). Namely, each instance in the dataset should contain accurate answers that are consistent with their supporting evidence, and conflicting answer pairs should be correctly identified. To that end, we randomly selected 40 instances from \ourdataset{} and hired a reviewer (a third worker) to validate the data. The reviewer, an undergraduate student, was trained for the task and paid a \$14 hourly wage (see \autoref{fig:app-qa-grounding-conflict} for the guidelines). The reviewer was instructed to assess whether each answer was supported by its linked evidence, and, independently, whether the pairs marked as conflicting were indeed conflicting.


In the collection phase, an answer was produced from a sentence within a passage. This means that, with respect to a passage, not all potential answers are necessarily included in the dataset. Therefore, given a passage, the correctness of an answer should be verified against the full passage, and not just any specific sentence within the passage. This matter is addressed in the evaluation metrics (\S\ref{subsec:evaluation-metrics}), and is also relevant for the validation phase of the dataset curation.


Following the guidelines, the reviewer labeled each answer in the 40 instances as ``correct'' or ``incorrect'' with respect to its linked evidence. Similarly, the reviewer labeled each conflicting pair in a binary fashion, validating whether the two answers conflict. Over this representing set of instances, we find a validation rate of 93\% for answer-correctness and 92\% for correctness of conflicting pairs. 
The meticulous dataset curation process, combined with the strong validation statistics, strongly suggest the high quality of \ourdataset{}.




\paragraph{Dataset properties.}
\repl{The validation phase above assesses the quality of individual instances in \ourtask{}. In addition, a}{A} high-level view of the dataset reveals properties that indicate the dataset's diversity and the challenges that it poses.

First, as indicated in \autoref{tab:dataset_comparison}, there are an average of 5.6 passages per question, and each passage has a length of 252 words, reflecting a realistic RAG setting with multiple long contexts.
Compared to other datasets for the MAQA task, \ourdataset{} features a relatively large number and length of passages per instance with high quality.
\repl{}{Additionally, \ourdataset{} includes 62 conflicting answer pairs each containing a mix of both conflicting and non-conflicting answers.
This subset proved challenging for models, as discussed in \S\ref{subsec:fine-grained-analysis}.}

\repl{First, as indicated in \autoref{tab:dataset_comparison}, there are an average of 5.6 passages per question, and each passage has a length of 252 words. This is suggestive of a RAG setting, where a response is generated based on several long passages. 
Compared to other datasets for the MAQA task, \ourdataset{} features a relatively large number and length of passages per instance, in addition to offering a significant improvement in overall quality.}{}

\repl{}{Subsequently, we observe that the number of evidence pieces linked to a single answer ranges from 1 to 17 $(\text{std}=2.7)$, with some answers appearing across multiple passages and others only once. In contrast, most existing datasets either do not specify evidence links \citep{xu2024debateqaevaluatingquestionanswering} or include far fewer per answer \citep{hou2024wikicontradictbenchmarkevaluatingllms, jiayang2024econdetectionresolutionevidence}.}
\repl{Subsequently, the dataset's diversity across various criteria enhances its complexity and makes it particularly challenging to solve. For example, we find that the number of evidence pieces linked to a single answer ranges from 1 to 17 $(\text{std}=2.7)$.
Specifically, some answers are considerably more salient as they appear across several given passages, while other answers are more anecdotal\repl{, appearing in one of many passages.}{} Meanwhile, the number of linked evidence pieces in most datasets is either unspecified \citep{xu2024debateqaevaluatingquestionanswering} or considerably smaller \citep{hou2024wikicontradictbenchmarkevaluatingllms, jiayang2024econdetectionresolutionevidence}(i.e., only a single evidence per answer).}{}
\repl{Next, the number of available passages per question ranges from 2 to 10 $(\text{std}=2.0)$, and their lengths range from 7 to 8043 $(\text{std}=522)$.
The diversity across the listed dimensions adds to the complexity that models must contend with when processing \ourdataset{}.}{See \autoref{app:data-statistics} for additional dataset statistics.}

\repl{Finally, ours is the first MAQA dataset that consists of questions with multiple answers that can conflict or support each other. Specifically, out of 89 WH-questions in our dataset that are accompanied by conflicting passages, 24 questions have answer pairs that conflict each other and also answer pairs that support each other.
This property puts models to test on a realistic scenario, where answers to the same question can both conflict and support each other.}{}

Taken together, the dataset's quality, diversity, and level of challenge make \ourdataset{} a valuable resource for studying model behavior in realistic settings.

%% file: 05experimental_setting.tex
\section{Experimental Setup}
\label{sec:experiment}

In this section, we outline our experimental setup for evaluating how well models detect and communicate conflicts.
We describe two prompting modes for the conflict-aware MAQA task (\S\ref{subsec:experiments-prompting-modes}) that will be applied on eight top‐performing LLMs (\S\ref{subsec:experiments-models}). Their performance will be assessed using two evaluation criteria adapted from prior work (\S\ref{subsec:evaluation-metrics}).


\subsection{LLM Prompting Modes}\label{subsec:experimental-setting}\label{subsec:experiments-prompting-modes}


To assess the performance of state‑of‑the‑art LLMs in a QA setting in which input passages may contain multiple, potentially conflicting answers, we conduct experiments using our dataset under two prompting modes --- defaultive and conflict-aware, similar to \citet{hou2024wikicontradictbenchmarkevaluatingllms}.

In MAQA and RAG settings, a system is expected to generate natural‐language responses that coherently articulate the information requested by an input instruction. We refer to this as the \textbf{defaultive} prompting mode, where the prompt simply states to answer a question based on the given sources.
In the \textbf{conflict-aware} prompting mode, the model is also explicitly instructed to identify and indicate any conflicts that arise when answering the question (full prompts are in \autoref{fig:app-dataset-prompts} in App. \ref{app:experimental-setting}).

\subsection{Tested Models}\label{subsec:experiments-models}

We selected eight top-performing open- and closed-source LLMs to evaluate in our setting. 
Specifically, since conflict-aware MAQA requires for strong reasoning abilities, to successfully identify conflicts across multiple passages, we selected four LLMs with an explicit reasoning step. 
We employed two flagship closed-source LLMs: \texttt{Gemini 2.5 Pro},\footnote{\url{https://blog.google/technology/google-deepmind/gemini-model-thinking-updates-march-2025}} and OpenAI's \texttt{o3},\footnote{\label{footnote:o3-o4}\url{https://openai.com/index/o3-o4-mini-system-card}} and to allow reproducibility, we selected two open-source reasoning language models: \texttt{DeepSeek‑R1} \cite{DBLP:journals/corr/abs-2501-12948} and \texttt{Qwen3-235B-A22B}.\footnote{\url{https://qwenlm.github.io/blog/qwen3}}
Finally, we selected four popular non-reasoning models: \texttt{gpt-4o} \cite{DBLP:journals/corr/abs-2410-21276}, \texttt{Gemini 2.0 Flash},\footnote{\url{https://blog.google/technology/google-deepmind/google-gemini-ai-update-december-2024}} \texttt{Qwen2.5-72B} \cite{DBLP:journals/corr/abs-2412-15115}, and \texttt{DeepSeek-V3} \cite{DBLP:journals/corr/abs-2412-19437}.
The eight models were evaluated on our \ourdataset{} dataset with the two prompting modes. 

\subsection{Evaluation Metrics}
\label{subsec:evaluation-metrics}

To evaluate model performance, we follow the two quality criteria of the task (\S\ref{sec:task-defintion}) --- \textit{answer quality} and \textit{conflict identification}.
To measure the two quality criteria, we define precision, recall, and $F_1$ measures per a conflict-aware MAQA instance, as described below. A system's overall scores are the average of each of the three instance-level metric scores.

\paragraph{Preparation step: decomposing the system response.} 

Consider an instance of conflict-aware MAQA (\S\ref{sec:task-defintion}), characterized by input question $q$ and passages $P$, and reference answers $A$ and conflicting answer pairs $C$. A system responds with a free text response $y$, which coherently addresses $q$ based on $P$. An interpreter (we use an \texttt{o4-mini} LLM) decomposes $y$ into a set of distinct answers $\hat{A}=\{\hat{a}_1, \dots, \hat{a}_k\}$, and then identifies all conflicting answer pairs in $y$ as $\hat{C}=\bigl\{\{\hat{a_i},\hat{a_j}\}\mid \hat{a_i}\text{ and }\hat{a_j}\text{ conflict within }y\bigr\}$. This decomposition step enables the evaluation, as described next.



\paragraph{Metrics for answer quality.}
To evaluate the correctness of an answer from the system's response, we adapt the recall and precision metrics from traditional MAQA tasks to our setting \cite{min-etal-2020-ambigqa,amouyal-etal-2023-qampari}.

$\mathrm{recall}_{\mathrm{ans}}$ is the fraction of reference answers $A$ found in the system response $y$, while $\mathrm{precision}_{\mathrm{ans}}$ is the fraction of system-derived answers $\hat{A}$ found in the given passages $P$. Formally:
\begin{align*}
(1)\quad &\mathrm{recall}_{\mathrm{ans}}
   \quad\ \ =\sum_{i = 1}^{m}
     \frac{\mathcal{J}_{\mathrm{ans}}(a_i, y)}{m}\\
(2)\quad &\mathrm{precision}_{\mathrm{ans}}
   = \sum_{i=1}^{k}
     \frac{\mathcal{J}_{\mathrm{ans}}(\hat{a}_i, P)}{k}
\end{align*}
where $\mathcal{J}_{\mathrm{ans}}(a_i,T)$ denotes a judge's binary decision for whether answer $a_i$ is found within the context $T$.
Accordingly, we define $F_{1_{\mathrm{ans}}}$ as the harmonic mean of $\mathrm{recall}_{\mathrm{ans}}$ and $\mathrm{precision}_{\mathrm{ans}}$.
For $\mathcal{J}_{\mathrm{ans}}$ we employ an \texttt{o4-mini\repl{-high}{}}\footnote{\url{https://openai.com/index/o3-o4-mini-system-card}\repl{}{ with high reasoning effort.}} LLM (LLM-as-a-judge \citealp{liu-etal-2023-g,DBLP:journals/corr/abs-2306-05685}), which shows \repl{moderate}{strong} correlations to human judgments (see \autoref{app:metrics-correlation} for details).

\paragraph{Metrics for conflict identification.} 

Prior works define conflict detection as a classification task: deciding whether a system‐generated answer signals the presence of conflicting information in the retrieved passages \cite{xu2024debateqaevaluatingquestionanswering, hou2024wikicontradictbenchmarkevaluatingllms}.
We extend their formulation to the general case where arbitrary pairs of distinct answers may conflict in the given passages.

$\mathrm{recall}_{\mathrm{conf}}$ measures the fraction of reference conflicts $C$ captured by the system-derived answers $\hat{A}$, while $\mathrm{precision}_{\mathrm{conf}}$ measures the fraction of system-derived conflicting answer pairs $\hat{C}$ that are also conflicting in passages $P$.\footnote{Since system‐derived answers may appear in the source passages, we evaluate precision for both criteria against these passages rather than relying on the reference answers.} Formally:
\begin{align*}
(3)\quad &\mathrm{recall}_{\mathrm{conf}}
   \quad\ \ =\sum_{\{a_i, a_j\}\in C}
     \frac{\mathcal{J}_{\mathrm{conf}}(a_i, a_j, y)}{\lvert C \rvert}\\
(4)\quad &\mathrm{precision}_{\mathrm{conf}}
   = \sum_{\{\hat{a}_i, \hat{a}_j\}\in \hat{C}}
     \frac{\mathcal{J}_{\mathrm{conf}}(\hat{a}_i, \hat{a}_j, P)}
          {\lvert \hat{C} \rvert}
\end{align*}
where $\mathcal{J}_{\mathrm{conf}}(a_i,a_j,T)$ denotes a judge's binary decision for whether $a_i$ and $a_j$ are indicated as conflicting in the context $T$.
Accordingly, we define $F_{1_{\mathrm{conf}}}$ as the harmonic mean of $\mathrm{recall}_{\mathrm{conf}}$ and $\mathrm{precision}_{\mathrm{conf}}$.
We use an LLM-as-a-judge for $\mathcal{J}_{\mathrm{conf}}$ as well.

%% file: 06Results.tex
\section{Results and Analysis}\label{sec:results}
\input{Tables/MainResults}

In this section, we first present results on the two subsets of \ourdataset{} under both prompting modes, and analyze the general trends (\S\ref{subsec:main-results}, \S\ref{subsec:non-conflict-results}). 
We then conduct a manual error analysis on a sample of model responses (\S\ref{subsec:error-analysis}), uncovering the techniques used by models to wrongly handle conflicts.

\subsection{Results on the \textit{Conflict} Subset}\label{subsec:main-results}

The performance of the eight models (\S\ref{subsec:experiments-models}) on the \textit{Conflict} subset of \ourdataset{} (\S\ref{sec:instance_structure}) is reported in \autoref{tab:main-results}. We compare the use of the defaultive prompt mode against the conflict-aware prompt mode~(\S\ref{subsec:experiments-prompting-modes}) based on the recall, precision and $F_1$ metric scores (\S\ref{subsec:evaluation-metrics})\repl{}{, and finally corroborate the observed trends with human judge}. 

\paragraph{Defaultive prompting.}



Under the default prompt, i.e., without any conflict-related guidance, the models exhibit relatively low \(\mathrm{recall}_{\mathrm{conf}}\) ($37.8-67.8$; 9th column in \autoref{tab:main-results}), indicating that they struggle to identify and convey conflicts without explicit instruction. 
Reasoning models generally outperform non‐reasoning ones --- for example, \texttt{DeepSeek‐V3} achieves only 55.5 \(\mathrm{recall}_{\mathrm{conf}}\), compared to 67.8 for \texttt{o3\repl{‐high}{}}.
Moreover, models appear to struggle less with retrieving answers from the passages, as indicated by the relatively higher $\mathrm{recall}_{\mathrm{ans}}$ scores. However there is still much room for improvement on this front as well.

In contrast, precision metrics remain uniformly high: both \(\mathrm{precision}_{\mathrm{conf}}\) and \(\mathrm{precision}_{\mathrm{ans}}\) exceed 80 and 92, respectively. This pattern is expected, since models are more prone to omission errors (which affect recall) than to producing irrelevant or spurious content.


\paragraph{Conflict-aware prompting.}

When explicitly prompted to identify conflicting answers (conflict‐aware mode), models performance improve significantly ($\Delta\mathrm{CA}$ columns). 
Notably, all eight models exhibit significant improvements in \(\mathrm{recall}_{\mathrm{conf}}\), ranging from 7.5 to 27.6 points, indicating that explicitly guiding models to seek conflicts is effective for identifying them.
Moreover, six of the eight models observed an increase in \(\mathrm{recall}_{\mathrm{ans}}\), while precision on both criteria remains more or less comparable. 
Overall, for conflict instances, applying conflict‐aware prompting is highly advantageous, improving both answer quality and conflict identification for most models. The subtle, yet meaningful, change in the prompt goes a long way for helping strong LLMs sense conflicts in the MAQA setting.

\input{Tables/OverallConflictSubsetsResults}\label{subsec:fine-grained-analysis}

\paragraph{\repl{}{Conflicting subsets analysis.}}\label{subsec:conflicting-subsets-analysis}

\repl{}{Next, we analyze model performance across three disjoint subsets of conflict instances in \ourdataset{}:
(1) \textbf{Yes/No} questions; 
(2) WH-questions in which all answer pairs are conflicting (\textbf{WH-conflict}); 
and (3) WH-questions that include both conflicting and non-conflicting answer pairs (\textbf{WH-mix}). 
\autoref{tab:overall-conflict-subsets-results} reports results averaged over eight models (see \autoref{fig:conflict_identification_all_subsets} in App. \ref{app:analysis-conflicts-types} for per model results). Notably, WH-mix is the most challenging subset and shows the smallest gains from conflict-aware prompting compared with the other two subsets. This suggests that the presence of both conflicting and non-conflicting signals within the same instance increases ambiguity, making it harder for models to reliably identify and reason about the conflicting information.}

\paragraph{\repl{}{Human judgment results.}}

\repl{}{To further corroborate the general trends observed above, we employed our evaluation protocol with a human judge to the two top-performing models on 120 \ourdataset{} instances (additional details in App. \ref{app:metrics-correlation}).
In \autoref{tab:human-main-results}, we report similar trends --- LLMs perform better on \(\mathrm{recall}_{\mathrm{ans}}\) and \(\mathrm{recall}_{\mathrm{conf}}\) when prompted in a conflict-aware setting (up by 5.3 and 33.3 points for the two models). Moreover, we measured a strong correlation (Pearson's $r > 0.62$; \autoref{tab:pearson_huma_llm_corr}) between the human and LLM judges for the four metrics from \S\ref{subsec:evaluation-metrics}, further supporting our findings.}

\subsection{Results on the \textit{No-conflict} Subset}\label{subsec:non-conflict-results}

Next, we conducted an experiment which mirrors the traditional MAQA task, requiring models to generate responses that incorporate all relevant answers from the passages, \textit{without} conflicting information.
Specifically, we test four models' performance on the \textit{No‐conflict} subset of \ourdataset{}, under both prompting modes, as a reference for the \textit{Conflict} subset's experiments above. 
All tested models (both reasoning and non-reasoning LLMs) exhibit high answer quality under defaultive prompting ($F_{1_\mathrm{ans}}>90$; see full results in \autoref{tab:support-results} in \autoref{app:results-support}).
This suggests that retrieving and integrating answers from passages is easier for models when no conflicts are present. However, when prompted in conflict‐aware mode, performance slightly degrades (up to 5.5 in $F_{1_\mathrm{ans}}$), possibly because the enforced knowledge of potential conflicts (even when none exist) somewhat disrupts the model's natural inference.

\subsection{Error Analysis -- Dealing with Conflicts}\label{subsec:error-analysis}

\input{Tables/RelatedWorkComparison}

The large performance gap between the \textit{Conflict} and \textit{No-conflict} subsets of \ourdataset{} (\S\ref{subsec:main-results}, \S\ref{subsec:non-conflict-results}) calls for further examination. To that end, we conducted a manual error analysis on 160 sampled system responses generated by four models under defaultive prompting, on the \textit{Conflict} subset (see \autoref{app:error-analysis-annotation} for full details). A human annotator categorized each system response into one of four main pre-defined error categories,\footnote{These categories were identified through a preliminary analysis for prominent error types, and inspired by \citet{jiayang2024econdetectionresolutionevidence}.} if an error was found (80 of the 160 instances), as follows:
\input{Tables/ErrorAnalysisBrief}


\noindent
See examples in \autoref{tab:error_types_with_examples} in the appendix. The four models exhibit similar distributions of error types.

When the examined models made mistakes in their responses, it generally seems as if they tried to overcome conflicts through manipulative techniques. In 42\% of the cases, they simply chose one answer in order to refrain from dealing with the conflicts. In 17\% of the cases, they generated a response that unified the conflicting answers into a general answer that does not disclose the conflicts (e.g., by averaging numbers). In about 13\% of the cases, the models fabricated information in an attempt to reconcile the conflicts. Finally, another approach was to simply respond with a general comment related to the question, without answering it.

These phenomena observed on high-end LLMs demonstrate the manners with which models attempt to overcome conflicting information. We call upon the research community to dive deeper into this matter, not only to solve conflict-related tasks such as ours, but also to better understand the way in which LLMs handle inconsistencies in knowledge.
Future research should explore developing systems that embrace the complexity of conflicts rather than simply resolving them.


%% file: Tables/MainResults.tex
\newcommand{\plusSig}[1]{\textcolor{pastelGreen}{$\twoheaduparrow#1$}}
\newcommand{\plus}[1]{\textcolor{pastelGreen}{$\uparrow#1$}}
\newcommand{\minus}[1]{\textcolor{pastelRed}{$\downarrow#1$}}
\newcommand{\minusSig}[1]{\textcolor{pastelRed}{$\twoheaddownarrow#1$}}

\begin{table*}[t]
\resizebox{\textwidth}{!}{%
\begin{tabular}{@{}cl|rrrrrr|rrrrrr@{}}
\toprule
\multicolumn{1}{l}{} & Model & \multicolumn{6}{c|}{Answer Quality} & \multicolumn{6}{c}{Conflict Identification} \\
\multicolumn{1}{l}{} & \multicolumn{1}{l|}{} & \multicolumn{2}{c}{$\mathrm{precision}_{\mathrm{ans}}$} & \multicolumn{2}{c}{$\mathrm{recall}_{\mathrm{ans}}$} & \multicolumn{2}{c|}{$\mathrm{F_1}_{\mathrm{ans}}$} & \multicolumn{2}{c}{$\mathrm{precision}_{\mathrm{conf}}$} & \multicolumn{2}{c}{$\mathrm{recall}_{\mathrm{conf}}$} & \multicolumn{2}{c}{$\mathrm{F_1}_{\mathrm{conf}}$} \\
\multicolumn{1}{l}{} & \multicolumn{1}{l|}{} & \multicolumn{1}{c}{D} & \multicolumn{1}{c}{$\Delta$CA} & \multicolumn{1}{c}{D} & \multicolumn{1}{c}{$\Delta$CA} & \multicolumn{1}{c}{D} & \multicolumn{1}{c|}{$\Delta$CA} & \multicolumn{1}{c}{D} & \multicolumn{1}{c}{$\Delta$CA} & \multicolumn{1}{c}{D} & \multicolumn{1}{c}{$\Delta$CA} & \multicolumn{1}{c}{D} & \multicolumn{1}{c}{$\Delta$CA} \\ \midrule
\multirow{4}{*}{\makecell[cl]{\textbf{\rotatebox[origin=c]{90}{non-}} \textbf{\rotatebox[origin=c]{90}{reasoning}}}} & \texttt{gpt-4o} & 92.4 & \plus{2.2} & 59.3 & \plusSig{6.7} & 67.5 & \plusSig{6.0} & 85.4 & \minus{2.2} & 37.8 & \plusSig{27.6} & 67.2 & \plus{12.5} \\
 & \texttt{Gemini 2.0 Flash} & 97.8 & \minus{1.0} & 63.8 & \plus{2.2} & 73.8 & \plus{0.7} & 82.5 & \plus{1.1} & 54.7 & \plusSig{7.5} & 66.4 & \plus{7.3} \\
 & \texttt{DeepSeek-V3} & 92.4 & \plus{1.1} & 65.2 & \plusSig{4.5} & 71.8 & \plusSig{4.5} & 83.0 & \minus{4.2} & 55.5 & \plusSig{15.4} & 75.8 & \plus{1.4} \\
 & \texttt{Qwen-2.5-72B} & 93.2 & \minus{0.1} & 60.6 & \plusSig{4.4} & 69.3 & \plus{3.0} & 80.2 & \minus{1.2} & 46.0 & \plusSig{24.7} & 72.2 & \plus{4.8} \\ \midrule
\multirow{4}{*}{\makecell[cl]{\textbf{\rotatebox[origin=c]{90}{reasoning}}}} & \texttt{o3\repl{-high}{}} & 93.8 & \minus{2.1} & 71.1 & \plusSig{7.3} & 77.0 & \plusSig{3.5} & 84.8 & \plus{1.9} & 67.8 & \plusSig{16.6} & 81.2 & \plus{6.9} \\
 & \texttt{Gemini 2.5 pro} & 96.2 & \minus{0.1} & 67.7 & \plus{3.6} & 75.6 & \plus{3.3} & 85.8 & \minus{4.6} & 61.8 & \plusSig{17.9} & 77.2 & \plus{3.0} \\
 & \texttt{DeepSeek-R1} & 95.3 & \minus{0.7} & 57.0 & \minus{0.2} & 64.6 & \plus{0.7} & 84.1 & \minus{6.8} & 50.6 & \plusSig{13.8} & 66.9 & \plus{3.4} \\
 & \texttt{Qwen-3-235B} & 94.3 & \minus{0.3} & 56.0 & \minus{3.3} & 62.8 & \minus{0.8} & 85.3 & \minus{1.5} & 50.3 & \plusSig{9.0} & 69.8 & \plus{0.8} \\ \bottomrule
\end{tabular}%
}
\caption{
Performance on the \textit{Conflict} subset of \ourdataset{} for non-reasoning (upper half) and reasoning models (lower half).
Average precision, recall, and $F_1$ scores are reported for the two quality criteria (\S\ref{subsec:evaluation-metrics}) --- answer quality (left section) and conflict identification (right section).
Results are shown for when models apply defaultive prompting (D), together with the absolute change in scores when applying conflict-aware prompting instead ($\Delta$CA).
Symbols \minus{} / \minusSig{} and \plus{} / \plusSig{} denote negative/significant-negative and positive/significant-positive changes. See \autoref{app:significant-tests} for details on significance‐testing.
Overall, conflict-aware prompting yields improvements for nearly all models across both evaluation criteria.
}
\label{tab:main-results}
\end{table*}

%% file: Tables/OverallConflictSubsetsResults.tex
\begin{table}[]
\resizebox{\columnwidth}{!}{%
\begin{tabular}{@{}lccc@{}}
\toprule
 & Yes/No & WH-All & WH-Mix \\ \midrule
defaultive     & $54.5 \pm 2.2$ & $53.1 \pm 3.6$ & $46.0 \pm 5.7$ \\
conflict-aware & $71.3 \pm 2.0$ & $72.9 \pm 3.2$ & $53.3 \pm 5.6$ \\ \bottomrule
\end{tabular}%
}
\caption{\repl{}{Average conflict-identification recall with 90\% confidence intervals across eight models, reported for both prompting modes and all three conflict subsets (as detailed in \S\ref{subsec:conflicting-subsets-analysis}).
}}
\label{tab:overall-conflict-subsets-results}
\end{table}

%% file: Tables/RelatedWorkComparison.tex

\begin{table*}[t]
\centering
\resizebox{\linewidth}{!}{%
\begin{tabular}{l l c c c c c c c }
\toprule
\textbf{Dataset} & \makecell[l]{\textbf{Collection} \\  \textbf{Method}} & \textbf{\#Instances} & \makecell{\textbf{Conflicting} \\ \textbf{Pairs}} & \makecell{\textbf{Conflict} \\ \textbf{Type}} & \makecell{\textbf{Avg.} \\ \textbf{\#Passages}} & \makecell{\textbf{Avg. Passage} \\ \textbf{Length (words)}} \\
\midrule
ConflictingQA \citep{wan2024evidencelanguagemodelsconvincing} & Automatic & 238 & All    & Factual & 9.2 & 314\\
WikiContradict \citep{hou2024wikicontradictbenchmarkevaluatingllms} & Manual &  253 & All & Factual & 2 & 43\\
DebateQA \citep{xu2024debateqaevaluatingquestionanswering} & Automatic &  2,941 & N.D.  & Point-of-View & 4.2 & 4687.6\\ 
ECon \citep{jiayang2024econdetectionresolutionevidence} & Automatic &  1,666 & All   & Factual & 3 & 47.3\\
\ourdataset{}~(Ours) & Manual & 677 & All \& Mixed \& None & Factual & 5.6 & 251.5 \\
\bottomrule
\end{tabular}
}
\caption{\repl{}{Representative} datasets for conflict‐aware QA. ``Conflicting Pairs'' indicates whether an instance in the dataset has only conflicting answers (\textit{All}), conflicting and non‐conflicting answers (\textit{Mixed}), no conflicting answers (\textit{None}), or whether that distinction is not well defined \textit{N.D.}. ``Avg.\ \# Passages'' denotes the average number of passages per instance.
}
\label{tab:dataset_comparison}
\end{table*}


%% file: Tables/ErrorAnalysisBrief.tex
\begin{table}[H]
\resizebox{\columnwidth}{!}{%
\begin{tabular}{@{}llr@{}}
\toprule
\textbf{Error} & \textbf{Description of response} & \multicolumn{1}{l}{\textbf{Frequency}} \\ \midrule
Choose & contains one reference answer & 42\% \\
Generalize & unifies answers by generalizing & 17\% \\
Resolve & hallucinates info to settle conflicts & 13\% \\
Refrain & does not answer question & 5\% \\ \bottomrule
\end{tabular}%
}
\label{tab:error-analysis-brief}
\end{table}

%% file: 02related_work_arie.tex
\section{Related Work}
\label{sec:related_work}

In multi-answer QA, a question may have multiple valid answers, each supported by its own evidence \citep{voorhees2004trec2003}. 
Although most datasets for this task generally assume that the different answers are consistent and complementary~\citep{kwiatkowski-etal-2019-natural,zhu-etal-2020-question,li-etal-2022-multispanqa,zhong2022romqabenchmarkrobustmultievidence,amouyal-etal-2023-qampari}, in real-world scenarios, a query can expose conflicts or discrepancies between the different textual sources. 

Yet, there are only \repl{a handful of}{several} QA datasets that address conflicting answers, each exhibiting its own set of limitations. \autoref{tab:dataset_comparison} shows the differences between \ourdataset{} and existing benchmarks. 
In general, conflicting QA instances in naturally occurring texts are scarce, hence a popular strategy of prior works is to automatically introduce synthetic misinformation in texts,  generating this way conflicting evidences for a QA instance \cite{jiayang2024econdetectionresolutionevidence,liu2025opendomainquestionanswering,su2024conflictbankbenchmarkevaluatinginfluence, wang2025retrievalaugmentedgenerationconflictingevidence, ming2025faitheval}. This approach inherently introduces artificial biases for the types of conflicts included in the dataset, 
as determined by the synthetic generation method.
Another approach to derive conflicting instances involves utilizing existing Yes/No questions coupled with documents retrieved from search results that contain conflicting information \cite{wan2024evidencelanguagemodelsconvincing}. 

\repl{}{Other recent works focus on various types of conflicts in RAG settings, where there is a single correct answer \cite{liu-etal-2025-open}, multiple points of view \cite{xu2024debateqaevaluatingquestionanswering}, or a mix of different conflict types, such as temporal, misinformation, or opinion \cite{cattan2025draggedconflictsdetectingaddressing}.}
Most similar to our work, WikiContradict~\citep{hou2024wikicontradictbenchmarkevaluatingllms} includes human-annotated QAs that incorporate naturally-occurring (rather than synthetic) conflicting answers, found in  Wikipedia articles. Yet, the instances in this dataset are limited to only two relatively short evidence passages, which always contradict each other.


In contrast to existing resources, \ourdataset{} is a human-annotated dataset composed of naturally occurring conflicts between the different answers, covering both Yes/No and WH- questions. Additionally, each instance includes on average 5.5 passages. Importantly, our work is the first to collect fine-grained annotations for each pair of answers, where some answer pairs are conflicting while others are not (the ``mixed'' category in \autoref{tab:dataset_comparison}. This annotation scheme enables more realistic assessment of models' ability to identify naturally occurring conflicting answers, while distinguishing them from non-conflicting answers.

%% file: 07conclusions.tex
\section{Conclusion}

In this work we enhance the \textit{Conflict-Aware Multi-Answer QA} task by explicitly requiring conflict identification among answers. We create a dataset for the enhanced task, via a novel cost-effective methodology that leverages fact-checking datasets. Our \ourdataset{} dataset is a realistic, conflict‐rich benchmark that challenges current strong models. We test several state-of-the-art LLMs on the dataset, and show that models still struggle with surfacing conflicting answers consistently\repl{}{, particularly in instances that contain both conflicting and non-conflicting answers}, even when expressly prompted to be on the watch for potential conflicts. Finally, an error analysis of model responses exposes manners in which LLMs mishandle conflicts.


%% file: 10limitations.tex
\section*{Limitations}

We employed LLMs for many tasks throughout this paper, including conflict-aware MAQA, evaluation of several criteria, and response decomposition. While we conducted some reasonable prompt-engineering for these assignments, it is possible that even more effective prompts would improve or change results.

Since pre-trained LLMs' training datasets are not fully documented, we can’t rule out overlap with the underlying data used for creating our dataset, raising the risk of contamination.




%% file: 11Acknowledgements.tex

%% file: 90Appendix.tex
\appendix




\section{Fact‐Checking Datasets for Conflict-aware MAQA}
\label{sec:fact-checking-datasets}

We build upon two established fact‐checking benchmarks, \climatefever{} and \healthver{}, which provide real‐world claims paired with supporting, refuting, and neutral evidence passages.

\paragraph{\climatefever{}.}  
The \climatefever{} dataset \citep{diggelmann2021climatefeverdatasetverificationrealworld} includes 808 claims. Of these, 654 claims have only non‐conflicting evidence (all support), while 154 claims are labeled as \textit{disputed}, containing conflicting evidences.

\paragraph{\healthver{}.}  
The \healthver{} dataset \citep{sarrouti-etal-2021-evidence-based} comprises 1,084 claims. Among them, 607 claims feature only supporting evidence (no conflict), and 477 claims include both supporting and refuting evidence, yielding true conflicts.

To ensure a diverse and representative subset of questions, we address the high redundancy in \healthver{}, where many real-world health-related claims are duplicated. We randomly sampled a single instance per topic from the 477 conflicting claims spanning on 55 topic questions. This process resulted in a final subset of 55 unique instances. 

\paragraph{Licensing.}  
Neither \climatefever{} nor \healthver{} specify an explicit license. Upon publication, we will release \ourdataset{} under the CC BY 4.0 license,\footnote{\url{https://creativecommons.org/licenses/by/4.0/}} permitting unrestricted reuse with attribution for research purposes.

\section{Dataset Annotation Details}
\label{sec:annotation-tool}

\input{Figures/AnnotationTool}
\input{Figures/AppAnnotationGuidelines}

\paragraph{Annotator details.}
Our two annotators were undergraduate or graduate students, and are fluent English speakers. They underwent two training iterations, each on 10 instances. They were compensated at approximately \$14 per hour.
The annotators worked a combined total of 52 hours to prepare the \ourdataset{} dataset.
The annotators were informed that their annotations are for research purposes, and that they can terminate their participation in the process whenever they want.

\paragraph{Annotation tool.}
To facilitate the annotation process, we developed a dedicated annotation tool that supports question-writing, linking respective evidence, and labeling conflicting answer pairs.

We present screenshots of the annotation interface and guidelines for both WH and Yes/No questions in Figures~\ref{fig:qa_annotation_interface} and~\ref{fig:app-annotation-guidelines}, respectively. The provided instructions guided annotators in accurately identifying conflicts and extracting relevant answers based on the given evidence.
Additionally, annotators were instructed to record the evidence IDs that support each annotated answer. Our custom annotation tool further allows annotators to add as many answers as necessary and seamlessly author both WH‐type and binary (Yes/No) questions for each claim. While the guidelines place a strong emphasis on handling conflict instances, annotators were allowed to skip the conflict-oriented instructions when working on the support instances.

\section{Passage Extraction for Evidence Sentences}
\label{subsec:passage-extraction}

To simulate a realistic QA setting, we reverse the typical sentence‐level focus of fact‐checking datasets by retrieving entire passages surrounding each evidence span.

For \healthver{}, which is derived from scientific abstracts in the CORD-19 corpus\footnote{\url{https://www.semanticscholar.org/cord19}}, we locate the original abstract corresponding to each evidence sentence. We perform an exact string match of the sentence within the CORD-19 collection and extract the full abstract to serve as the passage context.

For \climatefever{}, which uses evidence drawn from Wikipedia, we scraped the English Wikipedia pages as of February~1,~2020 and converted them to plain text. We then employ the RapidFuzz fuzzy‐matching library\footnote{\url{https://github.com/maxbachmann/RapidFuzz}} to identify each evidence sentence within its article. Finally, we heuristically expand to the surrounding paragraph—defined by nearest blank lines or section headers—to create a coherent passage that preserves the original narrative flow.

This passage‐level extraction ensures that each QA instance reflects the broader context in which evidence appears, aligning our setup with realistic Retrieval‐Augmented Generation workflows.

\input{APPstatistics}

\input{91AppExperimentalSetting}

\input{Figures/dataset_correlation_guideline}

\input{92AppResults}

%% file: Figures/AnnotationTool.tex
\begin{figure*}[tb]
    \centering
    \includegraphics[width=1.0\textwidth]{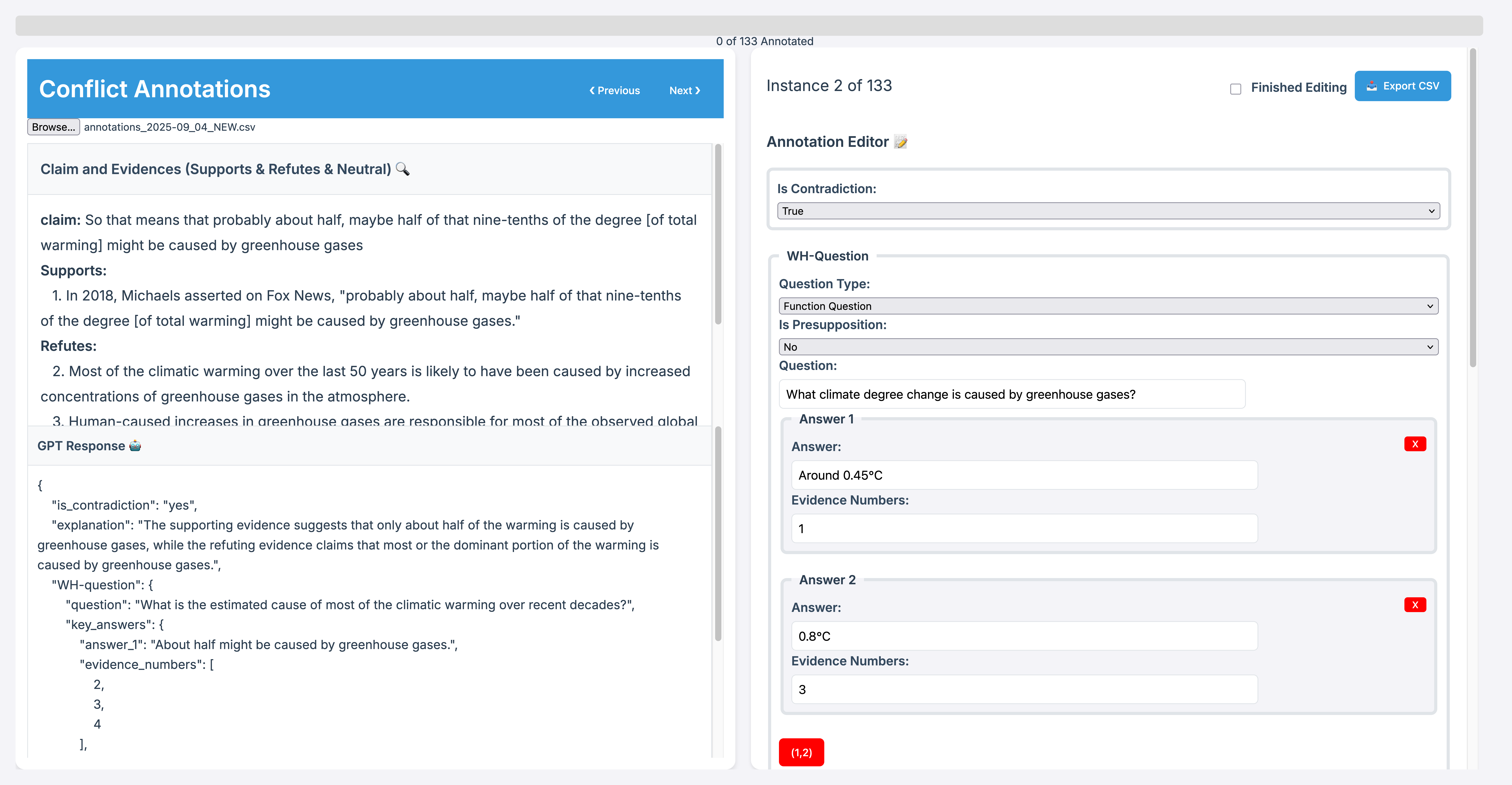}
    \caption{Screenshot of our annotation interface. On the left, annotators view the fact‐checking instance, including the claim and evidence sentences grouped by their initial labels (support, refute, neutral). At the bottom, LLM‐generated WH and yes/no question–answer suggestions are displayed \repl{}{(see \autoref{fig:suggestion-for-qa} for the prompts)}. On the right, annotators write or edit their own questions and corresponding answers.}
    \label{fig:qa_annotation_interface}
\end{figure*}

%% file: Figures/AppAnnotationGuidelines.tex
\begin{figure*}[tb]
\centering
\begin{tcolorbox}[
  colback=black!5!white,
  colframe=black,
  coltitle=white,
  title={\ourdataset{} Manual Annotation Guidelines}, 
  fonttitle=\bfseries,
  width=\textwidth,
  boxrule=0.6pt,
  arc=1.5mm                                                       
]
        \begin{tcolorbox}[title=1. Task Overview,
                          colback=gray!5!white,
                          colframe=gray!50!black,
                          arc=2mm,
                          boxrule=0.4pt]
        \begin{itemize}[leftmargin=*,label=\textbullet]
          \item \underline{Our goal} is to reveal conflicts in data through questions and answers.
          \item \underline{You will be provided with} a claim along with supporting, refuting, and neutral evidence.
          \item Your task involves two main objectives:
          \begin{enumerate}[label=(\arabic*),leftmargin=*]
            \item \textbf{Conflict detection}: Assess whether there is a conflict between the supporting and refuting evidence.\\[2pt]
                  \colorbox{green!25}{\textbf{Yes}}\;
                  \colorbox{red!25}{\textbf{No}}\;
                  \colorbox{gray!25}{\textbf{Uncertain}}
            \item \textbf{Q--A Generation}: Formulate questions that capture the \textcolor{blue}{conflict} based on the given claim and evidence. For each question, provide the differing answers along with the evidence IDs supporting each answer.
          \end{enumerate}
        \end{itemize}
        \end{tcolorbox}
        
        \vspace{0.8em}

        \begin{tcolorbox}[title=2. Rules for Effective Question Formulation,
                          colback=gray!5!white,
                          colframe=gray!50!black,
                          arc=2mm,
                          boxrule=0.4pt]
        You should follow the following rules when writing the questions:
        \begin{enumerate}[leftmargin=*]
          \item \textbf{Conflict-inducing} – The question should prompt a conflicting response.
          \item \textbf{Specificity} – Ensure the question targets detailed aspects of the text rather than general topics.
          \item \textbf{Assumption-free} – The question should be free from specific assumptions (from the text).
        \end{enumerate}
        \end{tcolorbox}
        
        \vspace{0.8em}
        
        \begin{tcolorbox}[title=3. Rules for Effective Answer Formulation,
                          colback=green!5!white,
                          colframe=green!50!black,
                          arc=2mm,
                          boxrule=0.4pt]
        You should follow the following rules when writing the answers:
        \begin{enumerate}[leftmargin=*]
          \item \textbf{Completeness} – Ensure that the answers address all the evidence provided.
          \item \textbf{Conciseness} – Keep the answers brief and to the point while maintaining clarity.
          \item \textbf{Relevance} – Ensure that the answers directly address the question asked.
          \item \textbf{Atomicity} – Each answer should contain only a single response. If multiple answers exist, they should be separated into distinct answers rather than combined.
        \end{enumerate}
        \end{tcolorbox}

\end{tcolorbox}
\caption{Summary of the annotation guidelines for creating our \ourdataset{} dataset. The full guidelines will be provided along with the complete resources.}
\label{fig:app-annotation-guidelines}
\end{figure*}

%% file: APPstatistics.tex
\section{\ourdataset{}Statistics}\label{app:data-statistics}

\input{Tables/our_data_statistics}
\autoref{tab:support_conflict_distribution_merged} presents statistics for \ourdataset{},\footnote{For randomly sampled instances from the dataset, see \autoref{tab:sample-instances}.} including number of questions, and number of passages, answers and conflicts per question. The dataset comprises two subsets, one with only supporting answers (titled \textit{No-conflict} in the table) and another with conflicting answers (titled \textit{Conflict}). For both types, there are WH-questions as well as Yes/No questions.

%% file: Tables/our_data_statistics.tex
\begin{table}[t]
  \centering
  \scriptsize
  \setlength{\tabcolsep}{6pt}  
  \begin{tabular}{lcc}
    \toprule
      \textbf{Statistic} & \textbf{\textit{No-conflict}} & \textbf{\textit{Conflict}} \\
    \midrule
      \# Instances                & 408  & 269 \\[2pt]
      Avg.\ \# Passages           & 6.7  & 5.6 \\[2pt]
      Avg.\ \# Answers (WH) [min–max]   & 3.8 [2–13] & 2.6 [2–10] \\[2pt]
      Avg.\ \# Conflict Pairs     & –    & 1.3 \\[2pt]
    \bottomrule
  \end{tabular}
\caption{
    Statistics for our \ourdataset{} dataset. The three average stats are per instance.
    \textit{Avg.\ \# Passages} is the mean number of passages per instance; 
    \textit{Avg.\ \# Answers (WH) [min–max]} reports the average number of answers per WH question and the corresponding range;
    \textit{Avg.\ \# Conflict Pairs} is the average number of conflicting answer pairs per instance (e.g., always 1 for Yes/No). 
}
  \label{tab:support_conflict_distribution_merged}
\end{table}

%% file: 91AppExperimentalSetting.tex
\section{Details for the Experimental Setup}
\label{app:experimental-setting}

\input{Figures/AppDatasetPrompts}
\input{Tables/AppModelsDetails}
\input{Figures/AppLLMasaJudgePrompts}

In this section, we provide technical details of the experimental setup defined in Section~\ref{subsec:experiments-models}.
All experiments were conducted between May 1 and May 19, 2025, using the OpenAI,\footnote{\url{https://platform.openai.com}} Google,\footnote{\url{https://aistudio.google.com}} and Together.ai\footnote{\url{https://together.ai}} APIs.
The exact version tags for all models utilized in this work are listed in \autoref{tab:app-model-details}.

For reproducibility, we set the temperature to 0 for all models with a maximum number of generated tokens (\texttt{max\_tokens}) at 512. 
For the OpenAI models (\texttt{o3} and \texttt{o4-mini}), we configure \texttt{reasoning\_effort} to \texttt{high}.
For the Gemini models, we set \texttt{thinking\_budget} to 1024 for \texttt{Gemini~2.5~Pro} and to 0 for \texttt{Gemini~2.0~Flash}.
The total cost of the experiments using the three LLM APIs was approximately \$500.

The prompts used for model evaluation are shown in \autoref{fig:app-dataset-prompts}, while those for the LLM-as-a-judge are presented in \autoref{fig:app-judge-prompts}.
There are five prompts in total for the LLM‐as‐a‐judge: one for answer decomposition (Prompt 1) and four for the binary decision functions defined in Section~\ref{subsec:evaluation-metrics} (Prompts 2–5), namely, $\mathcal{J}_{\mathrm{ans}}(a_i, y)$, $\mathcal{J}_{\mathrm{ans}}(\hat{a}_i, P)$, $\mathcal{J}_{\mathrm{conf}}(a_i, a_j, y)$, and $\mathcal{J}_{\mathrm{conf}}(\hat{a}_i, \hat{a}_j, P)$.

%% file: Figures/AppDatasetPrompts.tex
\begin{figure}[tb]
    \begin{tcolorbox}[colback=black!5!white, colframe=black, coltitle=white, title=\ourdataset{} Dataset Prompt Template, fonttitle=\bfseries
    ]
        Provide a concise, single-sentence answer that includes every distinct answer to the following question, based on the given passages from multiple sources. \uline{If any answers conflict, clearly indicate which ones are in conflict while remaining objective and neutral.} \\
    
        Question: \{question\} \\
        Passages: 
        Passage \#1: \\ \{Passage 1 text\} \\ \\
        Passage \#2: \\ \{Passage 2 text\} \\
        \dots
    \end{tcolorbox}
    \caption{The prompt template used to prepare inputs for the LLMs when testing on \ourdataset{}. This is the template for both modes (defaultive prompting mode and conflict‑aware prompting mode) as defined in \S\ref{subsec:experimental-setting}. The underlined sentence is omitted in the defaultive mode and included in the conflict‑aware mode.}
    \label{fig:app-dataset-prompts}
\end{figure}

%% file: Tables/AppModelsDetails.tex
\begin{table}[t]
\resizebox{\columnwidth}{!}{%
\begin{tabular}{@{}llc@{}}
\toprule
\multicolumn{1}{l}{\textbf{Model Name}} & \textbf{Model Tag} & \textbf{Reasoning} \\ \midrule
\texttt{GPT-4o} & \texttt{gpt-4o-2024-08-06} & \xmark \\
\texttt{Gemini 2.0 Flash} & \texttt{gemini-2.0-flash-001} & \xmark \\
\texttt{DeepSeek-V3} & \texttt{deepseek-ai/DeepSeek-V3} & \xmark \\
\texttt{Qwen-2.5-72B} & \texttt{Qwen/Qwen2.5-72B-Instruct-Turbo} & \xmark \\ \midrule
\texttt{o3\repl{-high}{}} & \texttt{o3-2025-04-16} & \cmark \\
\texttt{Gemini 2.5 Pro} & \texttt{gemini-2.5-pro-preview-03-25} & \cmark \\
\texttt{DeepSeek-R1} & \texttt{deepseek-ai/DeepSeek-R1} & \cmark \\
\texttt{Qwen-3-235B} & \texttt{Qwen/Qwen3-235B-A22B-fp8} & \cmark \\ \midrule
\texttt{o4-mini\repl{-high}{}} & \texttt{o4-mini-2025-04-16} & \cmark \\ \bottomrule
\end{tabular}%
}
\caption{Exact model version tags for the models used in this work.}
\label{tab:app-model-details}
\end{table}

%% file: Figures/AppLLMasaJudgePrompts.tex
\begin{figure*}[tb]
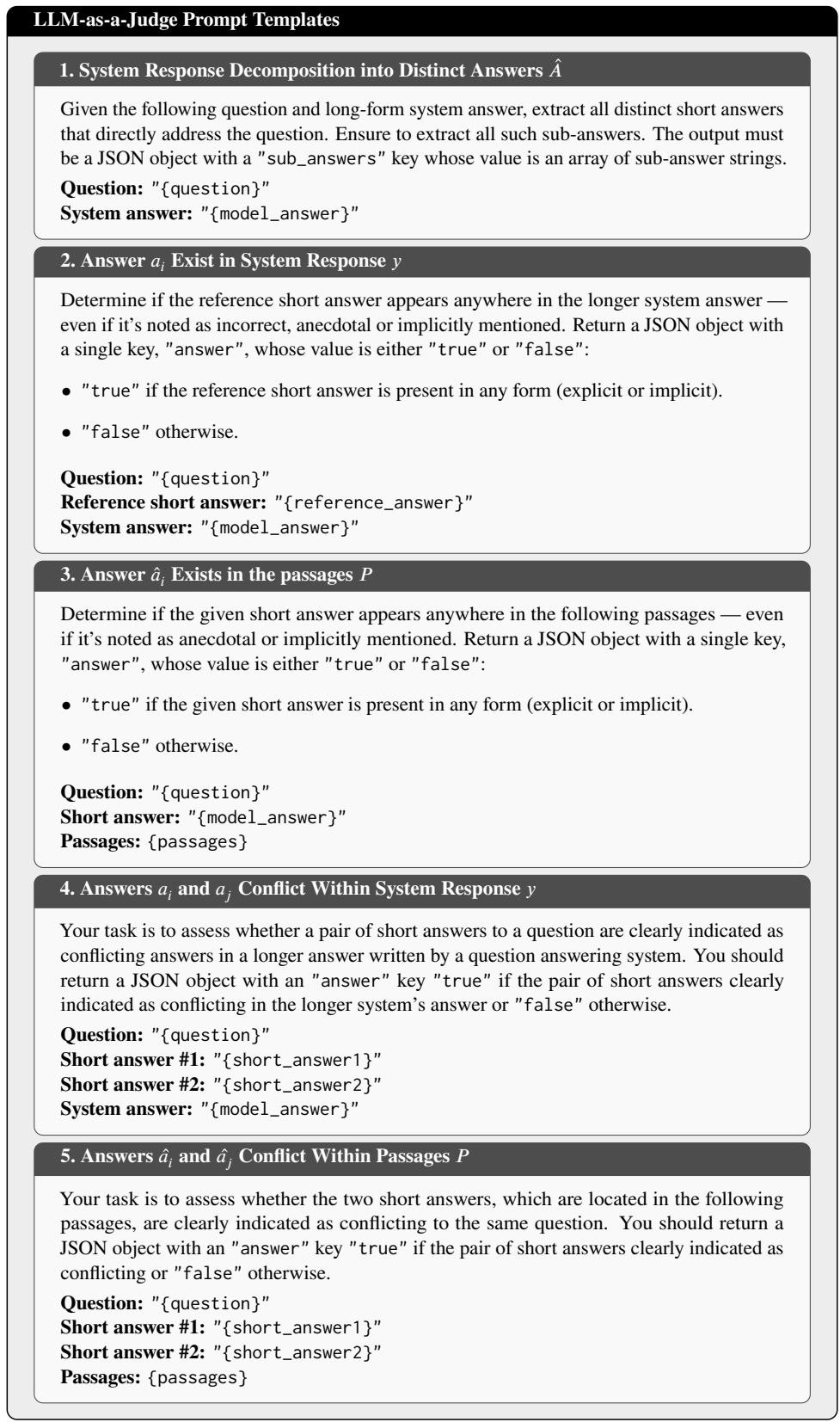

\centering
\resizebox{0.8\textwidth}{!}{%
\begin{tcolorbox}[
  colback=black!7!white,
  colframe=black,
  width=\textwidth,
  title=LLM-as-a-Judge Prompt Templates,
  fonttitle=\bfseries,
  boxrule=0.7pt,
  arc=1.8mm
]

\begin{tcolorbox}[title=1.~System Response Decomposition into Distinct Answers $\hat{A}$,
                  colback=gray!5!white,
                  colframe=gray!60!black,
                  fonttitle=\bfseries,
                  arc=1.6mm,
                  boxrule=0.5pt]
Given the following question and long-form system answer, extract all distinct short answers that directly address the question. Ensure to extract all such sub-answers.  
The output must be a JSON object with a \texttt{"sub\_answers"} key whose value is an array of sub-answer strings.

\smallskip
\textbf{Question:} \texttt{"\{question\}"} \\
\textbf{System answer:} \texttt{"\{model\_answer\}"}
\end{tcolorbox}

\vspace{-0.9em}

\begin{tcolorbox}[title=2.~Answer $a_i$ Exist in System Response $y$,
                  colback=gray!5!white,
                  colframe=gray!60!black,
                  fonttitle=\bfseries,
                  arc=1.6mm,
                  boxrule=0.5pt]
Determine if the reference short answer appears anywhere in the longer system answer --- even if it's noted as incorrect, anecdotal or implicitly mentioned.  
Return a JSON object with a single key, \texttt{"answer"}, whose value is either \texttt{"true"} or \texttt{"false"}:
\begin{itemize}[leftmargin=1em]
  \item \texttt{"true"} if the reference short answer is present in any form (explicit or implicit).
  \item \texttt{"false"} otherwise.
\end{itemize}

\smallskip
\textbf{Question:} \texttt{"\{question\}"} \\
\textbf{Reference short answer:} \texttt{"\{reference\_answer\}"} \\
\textbf{System answer:} \texttt{"\{model\_answer\}"}
\end{tcolorbox}

\vspace{-0.9em}

\begin{tcolorbox}[title=3.~Answer $\hat{a}_i$ Exists in the passages $P$,
                  colback=gray!5!white,
                  colframe=gray!60!black,
                  fonttitle=\bfseries,
                  arc=1.6mm,
                  boxrule=0.5pt]
Determine if the given short answer appears anywhere in the following passages --- even if it's noted as anecdotal or implicitly mentioned.  
Return a JSON object with a single key, \texttt{"answer"}, whose value is either \texttt{"true"} or \texttt{"false"}:
\begin{itemize}[leftmargin=1em]
  \item \texttt{"true"} if the given short answer is present in any form (explicit or implicit).
  \item \texttt{"false"} otherwise.
\end{itemize}

\smallskip
\textbf{Question:} \texttt{"\{question\}"} \\
\textbf{Short answer:} \texttt{"\{model\_answer\}"} \\
\textbf{Passages:} \texttt{\{passages\}}
\end{tcolorbox}

\vspace{-0.9em}

\begin{tcolorbox}[title=4.~Answers $a_i$ and $a_j$ Conflict Within System Response $y$,
                  colback=gray!5!white,
                  colframe=gray!60!black,
                  fonttitle=\bfseries,
                  arc=1.6mm,
                  boxrule=0.5pt]
Your task is to assess whether a pair of short answers to a question are clearly indicated as conflicting answers in a longer answer written by a question answering system.  
You should return a JSON object with an \texttt{"answer"} key \texttt{"true"} if the pair of short answers clearly indicated as conflicting in the longer system's answer or \texttt{"false"} otherwise.

\smallskip
\textbf{Question:} \texttt{"\{question\}"} \\
\textbf{Short answer \#1:} \texttt{"\{short\_answer1\}"} \\
\textbf{Short answer \#2:} \texttt{"\{short\_answer2\}"} \\
\textbf{System answer:} \texttt{"\{model\_answer\}"}
\end{tcolorbox}

\vspace{-0.9em}

\begin{tcolorbox}[title=5.~Answers $\hat{a_i}$ and $\hat{a_j}$ Conflict Within Passages $P$,
                  colback=gray!5!white,
                  colframe=gray!60!black,
                  fonttitle=\bfseries,
                  arc=1.6mm,
                  boxrule=0.5pt]
Your task is to assess whether the two short answers, which are located in the following passages, are clearly indicated as conflicting to the same question.  
You should return a JSON object with an \texttt{"answer"} key \texttt{"true"} if the pair of short answers clearly indicated as conflicting or \texttt{"false"} otherwise.

\smallskip
\textbf{Question:} \texttt{"\{question\}"} \\
\textbf{Short answer \#1:} \texttt{"\{short\_answer1\}"} \\
\textbf{Short answer \#2:} \texttt{"\{short\_answer2\}"} \\
\textbf{Passages:} \texttt{\{passages\}}
\end{tcolorbox}

\end{tcolorbox}
} 
\caption{Prompt templates used in our experiments for LLM-as-a-judge as detailed in Section~\ref{subsec:evaluation-metrics}.}
\label{fig:app-judge-prompts}
\end{figure*}

%% file: Figures/dataset_correlation_guideline.tex
\begin{figure*}[tb]
\centering
\resizebox{0.8\textwidth}{!}{%
\begin{tcolorbox}[
  colback=black!5!white,
  colframe=black,
  coltitle=white,
  title=Manual Annotation Guidelines for Correlation Assessment,
  fonttitle=\bfseries,
  width=\textwidth,
  boxrule=0.6pt,
  arc=1.5mm
]

\textbf{Task Overview:} You are given a question, human-annotated reference answers (referred to as \textit{reference\_answers}), human-annotated conflicting answer pairs (referred to as \textit{reference\_conflicting\_answer\_pairs}), associated passages, and a model-generated answer; please follow the steps below.

  \vspace{1em}

  \begin{tcolorbox}[title=1. Decompose the Model Answer into Atomic Answers,
                    colback=gray!5!white,
                    colframe=gray!50!black,
                    arc=2mm,
                    boxrule=0.4pt]
  \begin{itemize}[leftmargin=*,label=\textbullet]
    \item Read the \texttt{model\_answer} and split it into independent facts or claims (“atomic answers”).
    \item Number them 0, 1, 2, … in the order they appear.
    \item Enter them in the format:
    \begin{itemize}
      \item 0: First atomic answer…
      \item 1: Second atomic answer…
      \item 2: Third atomic answer…
    \end{itemize}
    \item $\rightarrow$ Fill in \texttt{Model Answers Decomposed}.
  \end{itemize}
  \end{tcolorbox}
  
  \vspace{0.8em}

  \begin{tcolorbox}[title=2. Match Decomposed Answers to Reference Answers,
                    colback=gray!5!white,
                    colframe=gray!50!black,
                    arc=2mm,
                    boxrule=0.4pt]
  \begin{itemize}[leftmargin=*,label=\textbullet]
    \item For each atomic answer (by index), check if it appears in the \texttt{reference\_answers}.
    \item List the indices of those that match, e.g., \texttt{[0, 2]}.
    \item $\rightarrow$ Fill in \texttt{Matched Answers in Reference Answers}.
  \end{itemize}
  \end{tcolorbox}

  \vspace{0.8em}

  \begin{tcolorbox}[title=3. Identify Conflicting Reference-Answer Pairs,
                    colback=gray!5!white,
                    colframe=gray!50!black,
                    arc=2mm,
                    boxrule=0.4pt]
  \begin{itemize}[leftmargin=*,label=\textbullet]
    \item Review the \texttt{reference\_conflicting\_answer\_pairs} and identify each pair that directly contradicts the other.
    \item In the model's response, look for contrastive cues like “however,” “but,” or “on the other hand.”
    \item Record each conflicting pair by their indices, e.g., \texttt{[(0,1), (1,2)]}.
    \item $\rightarrow$ Fill in \texttt{Conflicting Reference Answer Pairs Found}.
  \end{itemize}
  \end{tcolorbox}

  \vspace{0.8em}

  \begin{tcolorbox}[title=4. Match Model Answers to Passages or References,
                    colback=green!5!white,
                    colframe=green!50!black,
                    arc=2mm,
                    boxrule=0.4pt]
  \begin{itemize}[leftmargin=*,label=\textbullet]
    \item For each atomic answer, check if it is supported by either a \texttt{reference\_answer} or a paragraph.
    \item List the indices of atomic answers that are supported, e.g., \texttt{[0, 1]}.
    \item $\rightarrow$ Fill in \texttt{Found Model Answers in Reference Answers/Passages}.
  \end{itemize}
  \end{tcolorbox}

  \vspace{0.8em}

  \begin{tcolorbox}[title=5. Identify Conflicts among Model’s Answers,
                    colback=red!5!white,
                    colframe=red!50!black,
                    arc=2mm,
                    boxrule=0.4pt]
  \begin{itemize}[leftmargin=*,label=\textbullet]
    \item For each pair of atomic answers flagged by the model as conflicting, record:
    \begin{itemize}
      \item \texttt{(i, j): 1} if the conflict is correct.
      \item \texttt{(i, j): 0} if the conflict is incorrect.
    \end{itemize}
    \item Example: \texttt{(0, 2): 1, (0, 1): 0}
    \item $\rightarrow$ Fill in \texttt{Found Conflicting Model's Answers}.
  \end{itemize}
  \end{tcolorbox}

\end{tcolorbox}
} 
\caption{Annotation guidelines used by human annotators to assess the correlation between automatic LLM-based judgments and human evaluation. The process includes both conflict identification and answer quality metrics, supporting the measurement of recall and precision for each.}
\label{fig:app-decomposition-guidelines}
\end{figure*}

%% file: 92AppResults.tex
\section{\ourdataset{} \textit{No-conflict} Subset Results}\label{app:results-support}

\input{Tables/SupportResults}

In \autoref{tab:support-results}, we present results on the \textit{No-conflict} subset of \ourdataset{}, using the same metrics and notation conventions as in \autoref{tab:main-results}.

\section{Effectiveness of LLM-as-a-Judge}\label{app:metrics-correlation}
\repl{
Manually annotating answer quality and correct conflict detection is highly labor‐intensive. We employ a fast automatic judge --- an \texttt{o4-mini-high}\footnote{\url{https://openai.com/index/o3-o4-mini-system-card}} LLM --- which exhibits moderate correlation with human judgments.
Spearman's $\rho=0.44$ for $F_{1_\mathrm{ans}}$ ($p<0.01$) and $\rho=0.35$ for $F_{1_\mathrm{conf}}$ ($p<0.05$).
\eviatar{Change the order..}
To assess the correlation between automatic and human judgments, we randomly sampled 120 system responses from two models (\texttt{o3\repl{-high}{}} and \texttt{Gemini 2.5 Pro}) under both prompting modes (\S\ref{subsec:experiments-prompting-modes}).
A human judge then applied the judgment protocol explained in Section~\ref{subsec:evaluation-metrics}, with guidelines similar to the instructions in the prompts in \autoref{fig:app-judge-prompts}, decomposing answers and making the binary decisions that constitute both $F_{1_\mathrm{ans}}$ and $F_{1_\mathrm{conf}}$. Finally, we computed these $F_1$ scores and report the resulting correlation coefficients.
}{}

Manually annotating answer quality and conflict identification is expensive, so we rely on a fast automatic judge --- \texttt{o4-mini\repl{-high}{}}\footnote{\url{https://openai.com/index/o3-o4-mini-system-card}} --- and validate its agreement with humans. 
we randomly sampled 120 system responses from two models (\texttt{o3-high} and \texttt{Gemini 2.5 Pro}) under both prompting modes (\S\ref{subsec:experiments-prompting-modes}).
A human judge then applied the judgment protocol explained in Section~\ref{subsec:evaluation-metrics}, with guidelines similar to the instructions in the prompts in \autoref{fig:app-judge-prompts}, decomposing answers and making the binary decisions that constitute both metrics.

To assess the reliability of the automatic judge, we compute Pearson correlations between its labels and the human annotations (Table~\ref{tab:pearson_huma_llm_corr}). Both metrics show a strong positive correlation, with all correlation values exceeding 0.6.

Furthermore, the human evaluation results (\autoref{tab:human-main-results}) show performance trends similar to those from the automatic evaluation (\autoref{tab:main-results}). In both cases, conflict‐aware prompting leads to higher recall --- particularly for conflict identification.

\input{Tables/llms-as-judge-human-corr}
\input{Tables/human_main_results}

\section{Significance Testing on Results}\label{app:significant-tests}

To compute significance in \autoref{tab:main-results}, we conducted Wilcoxon signed-rank tests comparing the defaultive and conflict‐aware modes \cite{c4091bd3-d888-3152-8886-c284bf66a93a}.
We applied Pratt's conservative zero‐difference method \cite{Pratt01091959} and report significance at $p < 0.01$.

\section{Error Analysis Details}\label{app:error-analysis-annotation}
In order to assess model errors, we sampled 160 outputs, using the default prompt described in \S\ref{subsec:main-results}, from four models (\texttt{O3}, \texttt{DeepSeek-R1}, \texttt{GPT-4o}, and \texttt{DeepSeek-V3}). Annotation was performed by a graduate student at an hourly rate of \$14.
We sampled 40 instances for each model where 20 were Yes/No questions and 20 were WH-question.

\section{Conflict Subsets Analysis}\label{app:analysis-conflicts-types}
\autoref{fig:conflict_identification_all_subsets} compares conflict-identification recall across three subsets—Yes/No, WH-All (all answer pairs conflict), and WH-Mix (mixture of conflicting and non-conflicting pairs)-for each model under two prompting types: Defaultive and Contradict-Aware. Bars show mean recall per model. Two consistent trends emerge. First, Contradict-Aware prompting substantially improves performance for most models, especially on Yes/No and WH-All, while gains on WH-Mix are smaller. Second, WH-Mix is the hardest subset: it has the lowest base recall under Defaultive prompting and remains lowest even after Contradict-Aware prompting. This suggests that instances containing both conflicting and non-conflicting evidence introduce challenge that current models struggle to resolve.

\input{Figures/conflcit_identification_subsets}

\section{AI Assistance}
Throughout this project, we were assisted by AI tools to accelerate both code implementation (some code snippets) and manuscript preparation (local rephrasing). We carefully reviewed and refined all AI-generated content to ensure technical accuracy and stylistic consistency.

\input{Figures/error_categories}
\input{Figures/error_analysis_guideline}
\input{Figures/examples_from_data}

\input{Figures/qa_suggestion}




\input{Figures/data-quality-guidline}

%% file: Tables/SupportResults.tex
\begin{table}[t]
\resizebox{\columnwidth}{!}{%
\begin{tabular}{@{}>{\footnotesize}rl|rrrrrr@{}}
\toprule
\multicolumn{1}{l}{} & Model & \multicolumn{6}{c}{Answer Quality} \\
\multicolumn{1}{l}{} & \multicolumn{1}{l|}{} & \multicolumn{2}{c}{$\mathrm{precision}_{\mathrm{ans}}$} & \multicolumn{2}{c}{$\mathrm{recall}_{\mathrm{ans}}$} & \multicolumn{2}{c}{$\mathrm{F_1}_{\mathrm{ans}}$} \\ \addlinespace[5pt]
\multicolumn{1}{l}{} & \multicolumn{1}{l|}{} & \multicolumn{1}{c}{D} & \multicolumn{1}{c}{$\Delta$CA} & \multicolumn{1}{c}{D} & \multicolumn{1}{c}{$\Delta$CA} & \multicolumn{1}{c}{D} & \multicolumn{1}{c}{$\Delta$CA} \\ \addlinespace[5pt] \midrule \addlinespace[5pt]
\multirow{2}{*}{\makecell[cl]{\textbf{\rotatebox[origin=c]{90}{non-}} \textbf{\rotatebox[origin=c]{90}{reasoning}}}} & GPT-4o & 98.6 & \minus{1.4} & 95.6 & \plus{0.7} & 95.9 & \minus{0.1} \\  \addlinespace[5pt]
 & DeepSeek-V3 & 96.9 & \minus{1.1} & 95.5 & \plus{0.1} & 95.6 & \minus{1.0} \\  \addlinespace[5pt] \midrule \addlinespace[5pt]
\multirow{2}{*}{\makecell[cl]{\textbf{\rotatebox[origin=c]{90}{reasoning}}}} & o3\repl{-high}{} & 98.3 & \minus{1.5} & 97.2 & {0.0} & 97.5 & \minusSig{1.2} \\ \addlinespace[5pt]
 & DeepSeek-R1 & 97.6 & \minus{1.3} & 89.7 & \minus{4.2} & 90.7 & \minusSig{5.5} \\  \addlinespace[5pt] \bottomrule
\end{tabular}%
}
\caption{Average precision, recall, and $F_1$ percentages for the answer-quality criterion (\S\ref{subsec:evaluation-metrics}) on the \textit{No-conflict} subset of \ourdataset{}, reported for four LLMs. 
All models exhibit high scores (\(>89\)) on all measures.
See the caption of \autoref{tab:main-results} for further details on notations.}
\label{tab:support-results}
\end{table}

%% file: Tables/llms-as-judge-human-corr.tex
\begin{table}[t]
\centering
\resizebox{0.8\columnwidth}{!}{%
\begin{tabular}{@{}>{\footnotesize}l
    S[table-format=1.4]
    S[table-format=1.4]@{}}
\toprule
Evaluation Metric & {Recall} & {Precision} \\ \addlinespace[4pt]
\midrule \addlinespace[4pt]
Answer quality & 0.6645 & 0.6221 \\
Conflict identification & 0.6744 & 0.6371 \\
\bottomrule
\end{tabular}%
}
\caption{\repl{}{Pearson's $r$ correlation coefficients for recall and precision, computed between human and LLM-as-judge over 120 samples ($p < 0.00003$). The results show strong correlations for all metrics between human and LLM judgments.}}
\label{tab:pearson_huma_llm_corr}
\end{table}

%% file: Tables/human_main_results.tex


\begin{table*}[t]
\resizebox{\textwidth}{!}{%
\begin{tabular}{@{}cl|rrrrrr|rrrrrr@{}}
\toprule
\multicolumn{1}{l}{} & Model
& \multicolumn{6}{c|}{Answer Quality}
& \multicolumn{6}{c}{Conflict Identification} \\
\multicolumn{1}{l}{} & \multicolumn{1}{l|}{}
& \multicolumn{2}{c}{$\mathrm{precision}_{\mathrm{ans}}$}
& \multicolumn{2}{c}{$\mathrm{recall}_{\mathrm{ans}}$}
& \multicolumn{2}{c|}{$\mathrm{F_1}_{\mathrm{ans}}$}
& \multicolumn{2}{c}{$\mathrm{precision}_{\mathrm{conf}}$}
& \multicolumn{2}{c}{$\mathrm{recall}_{\mathrm{conf}}$}
& \multicolumn{2}{c}{$\mathrm{F_1}_{\mathrm{conf}}$} \\
\multicolumn{1}{l}{} & \multicolumn{1}{l|}{}
& \multicolumn{1}{c}{D} & \multicolumn{1}{c}{$\Delta$CA}
& \multicolumn{1}{c}{D} & \multicolumn{1}{c}{$\Delta$CA}
& \multicolumn{1}{c}{D} & \multicolumn{1}{c|}{$\Delta$CA}
& \multicolumn{1}{c}{D} & \multicolumn{1}{c}{$\Delta$CA}
& \multicolumn{1}{c}{D} & \multicolumn{1}{c}{$\Delta$CA}
& \multicolumn{1}{c}{D} & \multicolumn{1}{c}{$\Delta$CA} \\
\midrule
& \texttt{o3\repl{-high}{}}
& 85.5 & \minus{0.8}
& 84.8 & \plusSig{5.3}
& 85.1 & \plus{2.7}
& 100.0 & \minus{6.7}
& 46.7 & \plusSig{26.7}
& 85.7 & \plus{4.3} \\
& \texttt{gemini-2.5-pro}
& 79.9 & \minus{0.6}
& 83.4 & \plus{5.3}
& 81.6 & \plus{1.9}
& 81.8 & \plus{11.5}
& 46.7 & \plusSig{33.3}
& 57.1 & \plus{42.9} \\
\bottomrule
\end{tabular}%
}
\caption{
\repl{}{Human evaluation of performance on the \textit{Conflict} subset of \ourdataset{} for the two top performing models from \autoref{tab:main-results}, based on 120 instances.
Average precision, recall, and $F_1$ scores are reported (percent) for the two evaluation criteria (\S\ref{subsec:evaluation-metrics}): answer quality (left) and conflict identification (right).
Columns show results for default prompting (D) and the absolute change when using conflict-aware prompting ($\Delta$CA).
Symbols \minus{}/\minusSig{} and \plus{}/\plusSig{} denote negative/significant negative and positive/significant positive changes, respectively (using $p < 0.05$, see \autoref{app:significant-tests} for significance testing methodology).}
}

\label{tab:human-main-results}
\end{table*}

%% file: Figures/conflcit_identification_subsets.tex
\begin{figure*}[tb]
    \centering
    \includegraphics[width=1.0\textwidth]{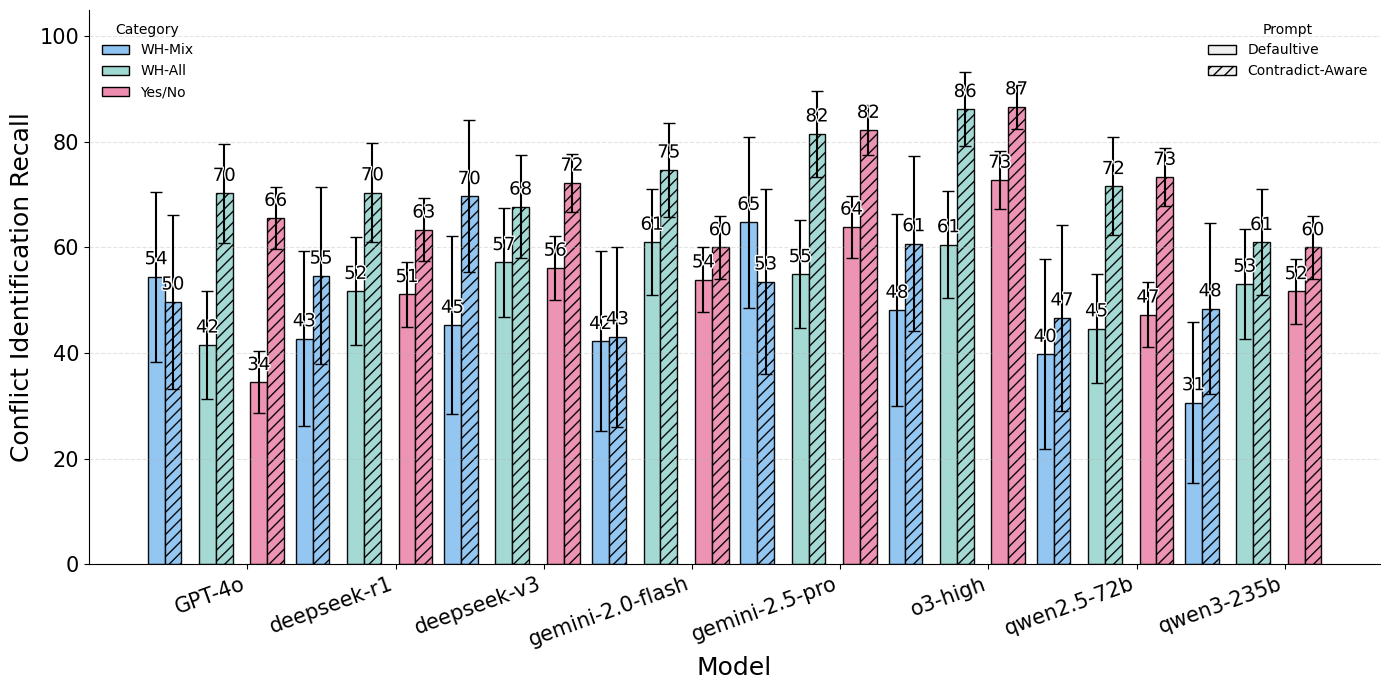}
    \caption{Conflict Identification Recall across models on three instance subsets (WH-Mix, WH-All, Yes/No), split by prompt type (defaultive vs. contradict-aware). Bars show mean conflict identification recall with 90\% confidence intervals. Overall, contradict-aware consistently improves recall relative to defaultive (\autoref{tab:main-results}); however, performance on WH-Mix is notably lower than on WH-All and Yes/No for nearly all models, and the gains from contradict-aware are smaller on this subset, highlighting the challenge when conflicting and non-conflicting signals co-occur within the same instance.}
    \label{fig:conflict_identification_all_subsets}
\end{figure*}

%% file: Figures/error_categories.tex
\begin{table*}[t]
\centering
\small
\renewcommand{\arraystretch}{1.5}  
\setlength{\tabcolsep}{8pt}        
\begin{tabular}{p{2.3cm}p{5cm}p{7cm}}
\toprule
\textbf{Error Type} & \textbf{Description} & \textbf{Example (Question / Reference answers / Model answer)} \\
\midrule

Choose Answer & 
The model outputs only one of the answers, omitting the others. &
\begin{minipage}[t]{\linewidth}\raggedright
\textbf{Q:} What is the rate of ice mass loss in Antarctica? \par
\smallskip
\textbf{Reference:} “+82 Gt/yr” and “–220 Gt/yr” \par
\smallskip
\textbf{Model:} “Antarctica is gaining 82 Gt of ice per year.”
\end{minipage} \\

\addlinespace

Answer Generalization & 
The model summarizes conflicting values vaguely (e.g., as an average or range) instead of presenting them distinctly. &
\begin{minipage}[t]{\linewidth}\raggedright
\textbf{Q:} How much has global temperature risen? \par
\smallskip
\textbf{Reference:} “0.45 °C” and “0.8 °C” \par
\smallskip
\textbf{Model:} “Temperatures have increased between 0.45 to 0.8.”
\end{minipage} \\

\addlinespace

Conflict Resolution & 
The model mentions all answers but presents them as if there is no conflict, possibly with hallucinatory information. &
\begin{minipage}[t]{\linewidth}\raggedright
\textbf{Q:} What is the estimated global temperature rise since 1900? \par
\smallskip
\textbf{Reference:} “0.45 °C” and “0.8 °C” \par
\smallskip
\textbf{Model:} “Until 1945, the rise was 0.45 °C, and then 0.8 °C.”
\end{minipage} \\

\addlinespace

Refrain from answering & 
The model does not provide an answer to the question or provides an irrelevant response. &
\begin{minipage}[t]{\linewidth}\raggedright
\textbf{Q:} What is the estimated rate of ice loss from Greenland per year? \par
\smallskip
\textbf{Reference:} “Between 200 and 300 Gt/yr” and “Approximately 220 Gt/yr” \par
\smallskip
\textbf{Model:} “Greenland is a large landmass covered in ice.”
\end{minipage} \\

\addlinespace

\bottomrule
\end{tabular}
\caption{Error types for the error analysis. Each instance is labeled with a single error type based on the model's ability to reflect, miss, or misrepresent the underlying conflict.}
\label{tab:error_types_with_examples}
\end{table*}

%% file: Figures/error_analysis_guideline.tex
\begin{figure*}[t]
\centering
\begin{tcolorbox}[
  colback=black!5!white,
  colframe=black,
  title={Manual Annotation Guidelines for Error Analysis},
  fonttitle=\bfseries,
  fontupper=\small, 
  boxsep=2pt,
  left=4pt, right=4pt, top=3pt, bottom=3pt,
  arc=1.5mm,
  width=\linewidth
]

\textbf{Task Overview:} You are provided with a question, the model’s answer for that question, and a set of annotated reference answers. First, review the list of error categories in \autoref{tab:error_types_with_examples}. Then, for each model output, classify it into the single most appropriate category, based on the explanation that best matches the model’s output. \\[0.5em]

\textbf{Steps:}
\begin{itemize}[leftmargin=1em, itemsep=2pt]
  \item Read the \textbf{question}, \textbf{reference answers}, and the \textbf{model’s answer}.
  \item If the model presents both sides with clear contrast (e.g., “but,” “however”), label it \texttt{-2}.
  \item Otherwise, choose a single error ID that best describes the model error.
  \item Use \texttt{--1} if the error does not fit any predefined category and describe it in the notes.
\end{itemize}

\textbf{Note:} Annotate each instance with only one ID. Include a brief justification if necessary.

\end{tcolorbox}
\caption{Annotation guidelines used by human annotators for conducting error analysis on outputs from different models.}
\label{fig:annotation-guideline}
\end{figure*}

%% file: Figures/examples_from_data.tex
\begin{table*}[ht]
  \centering
  \resizebox{\textwidth}{!}{%
    \begin{tabular}{@{} ll lp{6cm} l @{}}
      \toprule
      \textbf{Source Data} & \textbf{Q‐Type} & \textbf{A‐Type} & \textbf{Question} & \textbf{Answers} \\
      \midrule

      \climatefever{} & YN & No‐conflict & Do changes in land use, such as agriculture and deforestation, contribute to Earth's climate changes? 
      & 1. Yes \\

      \midrule

      \climatefever{} & YN & Conflict & Do greenhouse gases increase Earth's temperature? 
      & 1. Yes \\
      &    &          &                                              
      & 2. No \\

      \midrule

      \climatefever{} & wh & No‐conflict & What are the ranges of water vapor's contribution to the greenhouse effect? 
      & 1. Between 36\% and 66\% under clear sky conditions \\
      &    &            &                                                                                                   
      & 2. Between 66\% and 85\% when including clouds \\

      \midrule

      \climatefever{} & wh & Conflict & What is the perceived level of scientific agreement on climate change? 
      & 1. There is a scientific consensus on human-caused climate change \\
      &    &          &                                                                                             
      & 2. Scientific opinion is evenly divided or completely unsettled \\

      \midrule

      \healthver{}    & YN & No‐conflict & Does favipiravir decrease viral replication in COVID-19 patients? 
      & 1. Yes \\

      \midrule

      \healthver{}    & YN & Conflict & Do natural remedies help prevent you from getting infected with COVID-19? 
      & 1. Yes \\
      &    &          &                                                                                      
      & 2. No \\

      \midrule

      \healthver{}    & wh & No‐conflict & What are the clinical symptoms observed in patients with COVID-19? 
      & 1. Fever \\
      &    &            &                                                                                      
      & 2. Dry cough \\
      &    &            &                                                                                      
      & 3. Sore throat \\
      &    &            &                                                                                      
      & 4. Dyspnea \\
      &    &            &                                                                                      
      & 5. Fatigue \\
      &    &            &                                                                                      
      & 6. Myalgia \\
      &    &            &                                                                                      
      & 7. Headache \\
      &    &            &                                                                                      
      & 8. Loss of smell and taste \\

      \midrule

      \healthver{}    & wh & Conflict & What is possible in regards to cat to human coronavirus transmission? 
      & 1. Cats can transmit the virus to humans \\
      &    &          &                                                                                                
      & 2. Cats cannot transmit the virus to humans \\

      \bottomrule
    \end{tabular}%
  }
  \caption{Eight randomly sampled instances from \ourdataset{}. 
  Q‐Type: questoin type YN = yes/no; wh = WH‐questions. 
  A‐Type: answer type.
  When the dataset is published, it will also include the associated passages supporting each answer. 
  For simplicity, and since conflict instances in this sample contain only two answers, we do not include which specific answer pairs are in conflict.}
  \label{tab:sample-instances}
\end{table*}

%% file: Figures/qa_suggestion.tex
\begin{figure*}[tb]
\centering

\begin{tcolorbox}[
  colback=black!5!white,
  colframe=black,
  coltitle=white,
  title={Prompt for \ourdataset{} Question and Answers Suggestion for Conflicting Instances},
  fonttitle=\bfseries,
  width=\linewidth,
  boxrule=0.6pt,
  arc=1.5mm,
  fontupper=\small, 
  boxsep=2pt,
  left=4pt, right=4pt,
  top=2pt, bottom=2pt,
  parskip=1pt,
  before upper={\vspace{1pt}},
  after upper={\vspace{1pt}}
]

You are given a claim along with multiple pieces of evidence, categorized as supporting, refuting, or neutral in relation to the claim. Your task is to analyze the evidence and perform the following subtasks:

(1) \textbf{Determine Contradiction:} Identify whether there is a direct contradiction between the supporting and refuting evidence. Respond with \texttt{"yes"} if such a contradiction exists, or \texttt{"no"} if not.

(2) \textbf{Explain Contradiction (if applicable):} Briefly explain how the supporting and refuting evidence conflict.

(3) \textbf{Generate a WH-Question (if applicable):} Create a factoid-style WH-question based on the conflicting information. Avoid directly referencing the contradiction. Provide short answers and indicate the evidence numbers supporting each.

(4) \textbf{Generate a Yes/No Question:} Frame a yes/no question that reflects the core contradiction or claim, supported by evidence.

\textbf{Output Format:}
\begin{verbatim}
{
  "is_contradiction": "yes" or "no",
  "explanation": "...",
  "WH-question": {
    "question": "...",
    "key_answers": {
      "answer_1": "...",
      "evidence_numbers": [...],
      ...
    }
  },
  "Yes/no-question": {
    "question": "...",
    "Yes_answers": [...],
    "No_answers": [...]
  }
}
{instance}
\end{verbatim}

\end{tcolorbox}

\vspace{1em}

\begin{tcolorbox}[
  colback=black!5!white,
  colframe=black,
  coltitle=white,
  title={Prompt for \ourdataset{} Question and Answers Suggestion for Non-conflicting Instances},
  fonttitle=\bfseries,
  width=\linewidth,
  boxrule=0.6pt,
  arc=1.5mm,
  fontupper=\small, 
  boxsep=2pt,
  left=4pt, right=4pt,
  top=2pt, bottom=2pt,
  parskip=1pt,
  before upper={\vspace{1pt}},
  after upper={\vspace{1pt}}
]

You are given a claim along with multiple pieces of evidence, categorized as supporting or neutral. Your task is to:

(1) \textbf{Generate a WH-Question:} Create a concise WH-question based on the supporting evidence. Provide short answers and list the supporting evidence numbers.

(2) \textbf{Generate a Yes/No Question:} Frame a yes/no question grounded in the supporting evidence and reflecting the claim's context.

\textbf{Output Format:}
\begin{verbatim}
{
  "WH-question": {
    "question": "...",
    "key_answers": {
      "answer_1": "...",
      "evidence_numbers": [...],
      ...
    }
  },
  "Yes/no-question": {
    "question": "...",
    "Yes_answers": [...],
    "No_answers": [...]
  }
}
{instance}
\end{verbatim}

\end{tcolorbox}

\caption{Prompt templates used to generate suggested questions and answers for annotators in the \ourdataset{} annotation process. The first prompt is applied to instances containing conflicting evidence, while the second is used for non-conflicting evidence.}
\label{fig:suggestion-for-qa}
\end{figure*}

%% file: Figures/data-quality-guidline.tex
\begin{figure*}[tb]
\centering
\resizebox{\textwidth}{!}{%
\begin{tcolorbox}[
  colback=black!5!white,
  colframe=black,
  coltitle=white,
  title={Annotation Guidelines for \ourdataset{} Dataset Quality Assessment},
  fonttitle=\bfseries,
  width=\textwidth,
  boxrule=0.6pt,
  arc=1.5mm
]

  \begin{tcolorbox}[title=1. Task Overview,
                    colback=gray!5!white,
                    colframe=gray!50!black,
                    arc=2mm,
                    boxrule=0.4pt]
  \begin{itemize}[leftmargin=*,label=\textbullet]
    \item You will be provided with:
    \begin{itemize}
      \item A \textbf{question} prompting the model's response
      \item A list of \textbf{answers}, each associated with a unique ID
      \item A list of \textbf{evidence} that may support the answers
      \item A list of \textbf{answer pair labeled as conflicting}, specified by their answer IDs (if any)
    \end{itemize}
    \item Your task consists of two main steps:
    \begin{enumerate}[label=(\arabic*),leftmargin=*]
      \item \textbf{Answer Quality:} For each answer, determine whether it is clearly supported by at least one of the provided evidence.
            \begin{itemize}
              \item Mark \texttt{1} if it is grounded (i.e., the answer is directly supported by any evidence).
              \item Mark \texttt{0} if none of the evidence supports the answer.
            \end{itemize}
      \item \textbf{Conflict Detection:} For each answer pair labeled as conflicting, determine whether they conflicting each other based on the provided evidence.
            \begin{itemize}
              \item Mark \texttt{1} if the two answers clearly conflicting each other.
              \item Mark \texttt{0} if the answers are compatible or describe different aspects that can co-exist.
            \end{itemize}
    \end{enumerate}
  \end{itemize}
  \end{tcolorbox}

  \vspace{0.8em}

  \begin{tcolorbox}[title=2. Notes and Clarifications,
                    colback=blue!5!white,
                    colframe=blue!50!black,
                    arc=2mm,
                    boxrule=0.4pt]
  \begin{itemize}[leftmargin=*,label=\textbullet]
    \item Focus only on what is explicitly stated in the evidence and answers. Avoid making assumptions beyond the given text.
    \item If a pair appears borderline or ambiguous, lean toward \texttt{0} and leave a short note.
    \item Use the comments field to explain any unclear cases or edge scenarios you encounter.
  \end{itemize}
  \end{tcolorbox}

\end{tcolorbox}
}
\caption{Annotation guideline to evaluate the quality of \ourdataset{}. We assess both answer quality and conflict identification using an external annotator. The annotation procedure includes verifying whether answers are supported by evidence and whether identified answer pairs are truly in conflict. Results show high agreement, indicating the task is well-defined and the annotation protocol is reliable.}
\label{fig:app-qa-grounding-conflict}
\end{figure*}